%% file: main.tex
\documentclass[10pt,twocolumn,letterpaper]{article}

\usepackage{cvpr}              
\usepackage{multirow}

\usepackage[dvipsnames]{xcolor}
\definecolor{cvprblue}{rgb}{0.21,0.49,0.74}
\usepackage[pagebackref,breaklinks,colorlinks,citecolor=cvprblue]{hyperref}

\def\paperID{7031} 
\def\confName{CVPR}
\def\confYear{2024}

\title{Lookahead Exploration with Neural Radiance Representation for \\ Continuous Vision-Language Navigation}

\author{Zihan Wang$^{1,2}$, Xiangyang Li$^{1,2}$, Jiahao Yang$^{1,2}$, Yeqi Liu$^{1,2}$, Junjie Hu$^{3,4}$, Ming Jiang$^5$, Shuqiang Jiang$^{1,2}$\\
$^{1}$Institute of Computing Technology, Chinese Academy of Sciences, Beijing, 100190, China\\
$^{2}$University of Chinese Academy of Sciences, Beijing, 100049, China\\
$^{3}$Department of Computer Science, University of Wisconsin, Madison, WI, USA\\
$^{4}$Department of Biostatistics and Medical Informatics, University of Wisconsin, Madison, WI, USA\\ $^{5}$Department of Human-centered Computing, Indiana University, Indianapolis, IN, USA\\
\tt\small zihan.wang@vipl.ict.ac.cn, lixiangyang@ict.ac.cn, \{jiahao.yang, yeqi.liu\}@vipl.ict.ac.cn,\\ 
\tt\small junjie.hu@wisc.edu, mj200@iu.edu, sqjiang@ict.ac.cn
}

\begin{document}
\maketitle
\input{sec/0_abstract}

\input{sec/1_intro}

\input{sec/2_related_work}

\input{sec/3_method}
\input{sec/4_experiment}

\input{sec/5_conclusion}

{
\small
\bibliographystyle{ieeenat_fullname}
\bibliography{main}
}

\input{sec/X_suppl}

\end{document}

%% file: sec/0_abstract.tex
\begin{abstract}
Vision-and-language navigation (VLN) enables the agent to navigate to a remote location following the natural language instruction in 3D environments. At each navigation step, the agent selects from possible candidate locations and then makes the move. For better navigation planning, the lookahead exploration strategy aims to effectively evaluate the agent's next action by accurately anticipating the future environment of candidate locations.
To this end, some existing works predict RGB images for future environments, while this strategy suffers from image distortion and high computational cost. To address these issues, we propose the pre-trained hierarchical neural radiance representation model (HNR) to produce multi-level semantic features for future environments, which are more robust and efficient
than pixel-wise RGB reconstruction.
Furthermore, with the predicted future environmental representations, our lookahead VLN model is able to 
construct the navigable future path tree and select the optimal path via efficient parallel evaluation. Extensive experiments on the VLN-CE datasets confirm the effectiveness of our method. The code is available at
\href{https://github.com/MrZihan/HNR-VLN}{https://github.com/MrZihan/HNR-VLN}

\end{abstract}

%% file: sec/1_intro.tex
\section{Introduction}
\label{sec:intro}

Vision-and-language navigation (VLN) tasks~\cite{VLN_2018vision,2020_RXR,2020reverie,Krantz2020r2r-ce} 
 require an agent to understand
natural language instructions and move to the destination. In the continuous environment setting (VLN-CE)~\cite{Krantz2020r2r-ce}, the navigation agent is free to traverse any unobstructed location with low-level actions (\textit{i.e.}, turn left 15 degrees, turn right 15 degrees, or move forward 0.25 meters), similar to
some visual navigation tasks~\cite{zsx_ICCV21,zsx_ECCV22,zsx_CVPR23,wang2023generating,wang2024camp}. As a result, the agent is more prone to entering the visual blind area compared to the discrete environment setting with perfectly predefined navigable locations. This
phenomenon raises a challenge to accurately represent future environments with visual occlusions, leading to incorrect action decisions.


\begin{figure}
\noindent\begin{minipage}[h]{1\columnwidth}%
\begin{center}
\includegraphics[width=0.8\columnwidth]{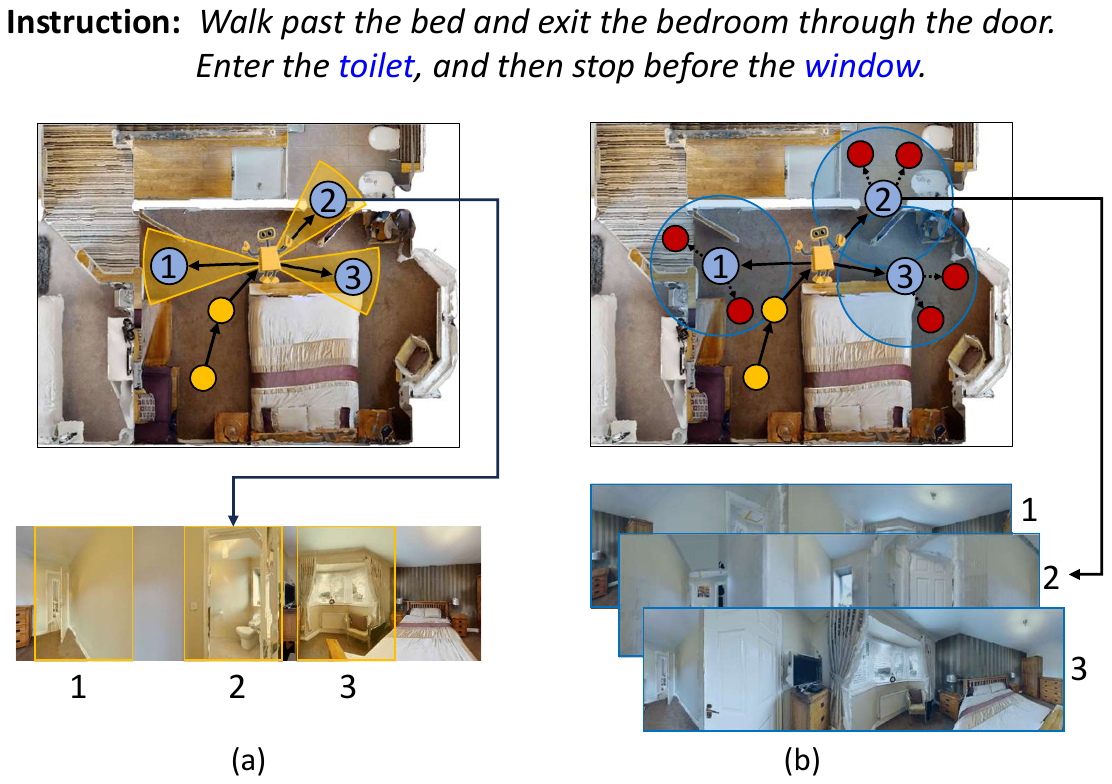}
\par\end{center}%
\end{minipage}
\vspace{-10pt}
\caption{Illustration of different methods to represent the navigable candidate locations. (a) uses the single-view observation (yellow sector area). (b) uses the panorama of the candidate location (blue circular area) to anticipate the future environment.}
\label{fig:introduction}
\vspace{-15pt}
\end{figure}

As illustrated in Figure~\ref{fig:introduction}(a), previous approaches~\cite{chen2021history,chen2022think-GL,Li2023KERM,li2023improving} mainly rely on single-view visual observation of the current location to perceive candidate locations, thus leading to a \textit{small restricted view}  (yellow sector area) due to the visual occlusions. In contrast, as shown in Figure~\ref{fig:introduction}(b), a \textit{comprehensive panorama} of the candidate location (blue circular area) is much more helpful to the agent for understanding the future environment and capturing critical visual cues in action decisions. \textit{Lookahead exploration}~\cite{feng2022uln,agarwal2019model} aims to enable an agent to explore steps forward before making a navigation decision. Unlike other \textit{lookahead} approaches, we adopt a strategy of exploring steps forward with environment anticipation, which helps current action decisions by predicting the future outcome of actions.

As generating future environment representations benefits much for action prediction, DREAMWALKER~\cite{wang2023dreamwalker} proposes to imagine panoramic images of the navigable candidates. With simulated images through the visual generation model~\cite{koh2021pathdreamer}, it 
gains promising performance. 
However, one major drawback of this method is the large distortions between the generated panoramic images and the actual images, introducing \textit{noisy} visual features to the agent. Due to the pixel-wise training objective, image generation model~\cite{koh2021pathdreamer} tends to fit the RGB values of local pixels, rather than focusing on key environmental semantics. Indeed, for unseen 3D environments, accurate RGB reconstruction is insurmountably difficult due to the high information redundancy of RGB images. In addition, generating high-resolution panoramic images is computing-intensive and time-consuming, 
increasing the delay of agent's responses. 


To anticipate future environments with higher quality and faster speed, we propose a pre-trained Hierarchical Neural Radiance (HNR) Representation Model that produces multi-level semantic representations of future candidate locations instead of generating panoramic images. Our semantic representations are learned through a vision-language embedding model (\textit{i.e.}, CLIP~\cite{radford2021learning}) that compresses the redundant information of RGB images and extracts the critical visual semantics associated with the language. Specifically at each step of navigation, the fine-grained grid features~\cite{wang2023gridmm} extracted by the CLIP model are saved into the feature cloud with 3D positions and orientations. To predict the semantic features of future environments, we sample points along camera rays, and aggregate features around these points to produce latent feature vectors and volume density, then use volume rendering techniques to composite these values into small-scale region features. Furthermore, multi-level encoders are adopted to produce multi-level semantic features for future environments, supervised by CLIP embeddings.

The advantages of our method over previous methods for future environment prediction are three-fold. First, our model directly predicts robust multi-level semantic features for future candidate locations, avoiding the difficulty of pixel-level image reconstruction in unseen environments as used in existing methods like RNR-Map~\cite{kwon2023renderable,taioli2023language} and DREAMWALKER~\cite{wang2023dreamwalker}. Second, as many empty regions in the future views can be caused by visual occlusions, we employ a hierarchical encoding method to predict features of these empty regions by integrating surrounding contexts at both region and view levels. 
Lastly, our method using volume rendering handles better spatial relationships in the 3D environment, which is challenging for 2D image generation methods~\cite{wang2023dreamwalker,li2023improving} that lack depth perception.

With the predicted high-quality future views of candidate locations, we propose a lookahead VLN model to evaluate the possible next actions. As shown in Figure~\ref{fig:introduction}(b), we predict more navigable locations (red nodes) in future environments and integrate them into a future path tree. The cross-modal graph encoder parallelly evaluates the matching score of different path branches in this path tree and selects the optimal candidate locations (blue nodes) to move.


In this work, our main contributions include: 
\begin{itemize}
\item We propose a hierarchical neural radiance representation model to produce multi-level semantic representations for future environments with better quality and efficiency. 

\item Utilizing predicted representations of future environments, we propose a lookahead VLN model to parallelly evaluate possible future paths in the path tree, thus improving navigation planning.

\item Extensive experiments demonstrate the effectiveness of our methods over existing methods in continuous vision-and-language navigation tasks.
\end{itemize}

%% file: sec/2_related_work.tex
\section{Related Work}
\label{sec:related_work}
\begin{figure*}[htbp]
\makebox[\textwidth][c]
{\includegraphics[width=0.75\paperwidth]{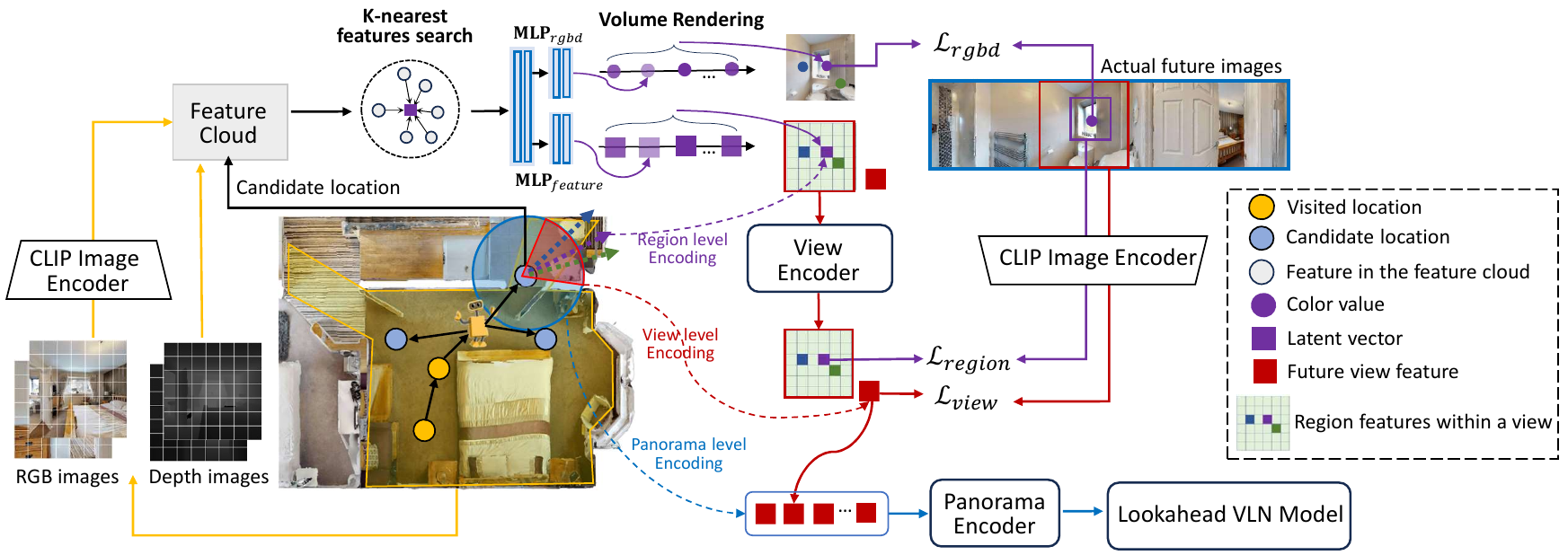}}
\vspace{-20pt}
\caption{The framework of the hierarchical neural radiance representation model (HNR). The HNR model encodes the observed environments (yellow area) into the feature cloud. Through aggregating k-nearest features, the MLP network predicts the latent vector and volume density of sampled points along the rendered ray. A region-level representation is encoded by compositing these latent vectors via volume rendering, then a view encoder is used to encode all region-level representations within a future view (red area) and obtain an entire future view representation. All future views of the candidate location can be combined as a panorama (blue area) to support navigation.}
\label{fig:framework}
\vspace{-15pt}
\end{figure*}

\noindent \textbf{Vision-and-Language Navigation (VLN).} VLN~\cite{VLN_2018vision,2020reverie,2020_RXR,Krantz2020r2r-ce} has received significant attention in recent years. The VLN tasks include step-by-step instructions such as R2R~\cite{VLN_2018vision} and RxR~\cite{2020_RXR}, navigation with dialogs such as CVDN~\cite{thomason2020cvdn}, and navigation for remote object grounding such as REVERIE~\cite{2020reverie} and SOON~\cite{zhu2021soon}. VLN-CE tasks convert the topologically-defined VLN tasks into continuous environments, such as R2R-CE~\cite{Krantz2020r2r-ce} and RxR-CE~\cite{2020_RXR}. Lots of previous methods focus on the representations of the visited environment during navigation. Among them, the recurrent unit~\cite{VLN_2018vision,2018-speaker,tan2019learning,2019reinforced,hong2021vln-bert}, explicitly encoded history sequence~\cite{pashevich2021episodic,chen2021history}, topological map~\cite{chen2022think-GL,an2023etpnav}, top-down semantic map~\cite{georgakis2022cm2,chen2022weakly,huang23vlmaps} and grid-based map~\cite{an2023bevbert,wang2023gridmm,liu2023bird} are usually adopted to represent the visited environment.

Although the representation of the visited environment in VLN has received continuous attention, future environmental representation and lookahead strategy have not been fully explored. VLN-SIG~\cite{li2023improving} generates
the semantics of future navigation views using visual codebook~\cite{ramesh2021zero} for better decision-making. DREAMWALKER~\cite{wang2023dreamwalker} utilizes an image generation model~\cite{koh2021pathdreamer} to generate panoramic images of future environments and predict future actions.  ULN~\cite{feng2022uln} and Active Exploration~\cite{wang2020active} explore steps forward for robust navigation decision-making, and Tactical Rewind~\cite{ke2019tactical} performs tree search for planning. In contrast to them, we propose a hierarchical neural radiance representation model to anticipate future environments and construct the navigable future path tree for long-term planning.

\noindent \textbf{Navigation with Neural Radiance Fields.} The neural radiance field (NeRF)~\cite{mildenhall2021nerf} has gained significant popularity in various AI tasks.
NeRF predicts the RGB color and density of a sampled point in a scene so that an image from an arbitrary viewpoint can be rendered.
However, the traditional NeRF methods with implicit MLP network can only synthesize view images in seen scenes, which makes it difficult to generalize to the unseen scenes and adapt to many embodied AI tasks.
To address this issue, GSN~\cite{devries2021unconstrained} proposes locally conditioned radiance fields, which encode observed images into a latent grid map, supporting rendering images in unseen environments. 
Based on GSN~\cite{devries2021unconstrained}, RNR-Map~\cite{kwon2023renderable} designs a localization framework using renderable neural radiance map for visual navigation, and Le-RNR-Map~\cite{taioli2023language} embeds CLIP features into RNR-Map for query search with natural language. RNR-Map and Le-RNR-Map all encode features into a 2D latent grid map, which loses a lot of 3D geometric detail.  The 3D feature cloud and hierarchical encoding in HNR provide better capabilities of spatial representation and multi-level context integration.

%% file: sec/3_method.tex
\section{Method}
\label{sec:method}

\vspace{-5pt}
\subsection{Navigation Setups}
\vspace{-5pt}
The HNR model focuses on the VLN-CE~\cite{Krantz2020r2r-ce,2020_RXR} tasks, where the agent navigates with low-level actions. 
Initialized at a starting location and given natural language instructions $\mathcal{W}$, the agent needs to explore the environment and reach the target location.
At time step $t$, the agent
observes panoramic RGB images $\mathcal{R}_{t}=\{r_{t,i}\}_{i=1}^{12}$ and the depth images $\mathcal{D}_{t}=\{d_{t,i}\}_{i=1}^{12}$ surrounding its current location (\textit{i.e.}, 12 view images with 30 degrees separation). 

During navigation, the agent's visual observations are encoded and stored into the feature cloud. Meanwhile, a pre-trained waypoint predictor~\cite{Hong2022bridging} is used to predict navigable candidates.
For each navigable candidate, the HNR model predicts 12 future view representations using hierarchical encoding method. Firstly, HNR uses the volume rendering method to aggregate the features from the feature cloud and produce region-level embeddings. Secondly, region-level embeddings within the same future view are fed into the view encoder and obtain the entire future view representation. These future representations of candidates help predict more and farther navigable locations (\textit{i.e.,} lookahead nodes). 
By integrating lookahead node representations into a future path tree, the lookahead VLN model evaluates the match scores of different path branches and selects the optimal candidate location.
Finally, a control module~\cite{an2023etpnav} is used to produce low-level actions to reach the selected candidate location.


\subsection{Hierarchical Neural Radiance Representation}\label{sec:HNR}
\subsubsection{Feature Cloud Encoding}\label{sec:feature_cloud}
To encode the observed visual information, as shown in Figure~\ref{fig:framework}, our HNR model stores the fine-grained visual features and their corresponding spatial information into the feature cloud $\mathcal{M}$. Specifically, at each navigation step $t$, for 12 observed RGB images $\mathcal{R}_{t}=\{r_{t,i}\}_{i=1}^{12}$, a pre-trained CLIP-ViT-B/32~\cite{radford2021learning} model is used to extract grid features $\{\textbf{g}_{t,i}\in\mathcal{\mathbb{R}}^{H\times W\times D}\}_{i=1}^{12}$.
For convenience, all the subscripts $(i,h,w)$ are denoted as $j$, where $j$ ranges from 1 to $J$, and $J=12 \cdot$$H$$\cdot$$W$. 
Through the downsized depth images $\{\textbf{d}_{t,i}\in\mathcal{\mathbb{R}}^{H\times W}\}_{i=1}^{12}$,
each grid feature $\textbf{g}_{t,j}\in {\mathbb{R}}^{D}$ is mapped to its 3D world position $P_{t,j}=[p_x,p_y,p_z]$ using camera pose $[\mathbf{R}, \mathbf{T}]$ and camera intrinsics $\mathbf{K}$ as follows: 

\vspace{-10pt}
\begin{align} 
\begin{split}
P_{t,j}
&= 
\begin{bmatrix}\textbf{d}_{t,j} \mathbf{R}^{-1} \mathbf{K}^{-1} 
\begin{bmatrix}
h \\
w \\
1 \\
\end{bmatrix} 
- \mathbf{T} \;
\end{bmatrix}^T
\end{split}\label{formula:inverse_projection}\end{align}

\vspace{-8pt}
To better represent the spatial information of features in the feature cloud, we introduce the horizontal orientation $\theta_{t,j}$ and the size $s_{t,j}$ of each observed grid region: 

\vspace{-10pt}
\begin{align}
s_{t,j}=1/W \cdot [\tan(\Theta_{HFOV}/2)\cdot \textbf{d}_{t,j}]
\end{align}

\vspace{-8pt}
\noindent where $W$ is the width of the feature map extracted by the CLIP model for each view, $\Theta_{HFOV}$ is the camera's horizontal field-of-view. All these grid features and their spatial information are stored into the feature cloud $\mathcal{M}$:
\begin{align}
\mathcal{M}_{t}= \mathcal{M}_{t-1} \cup \{[\textbf{g}_{t,j}, P_{t,j},\theta_{t,j},s_{t,j}]\}_{j=1}^{J}
\end{align}

\begin{figure}
\noindent\begin{minipage}[b]{1\columnwidth}%
\begin{center}
\includegraphics[width=0.9\columnwidth]{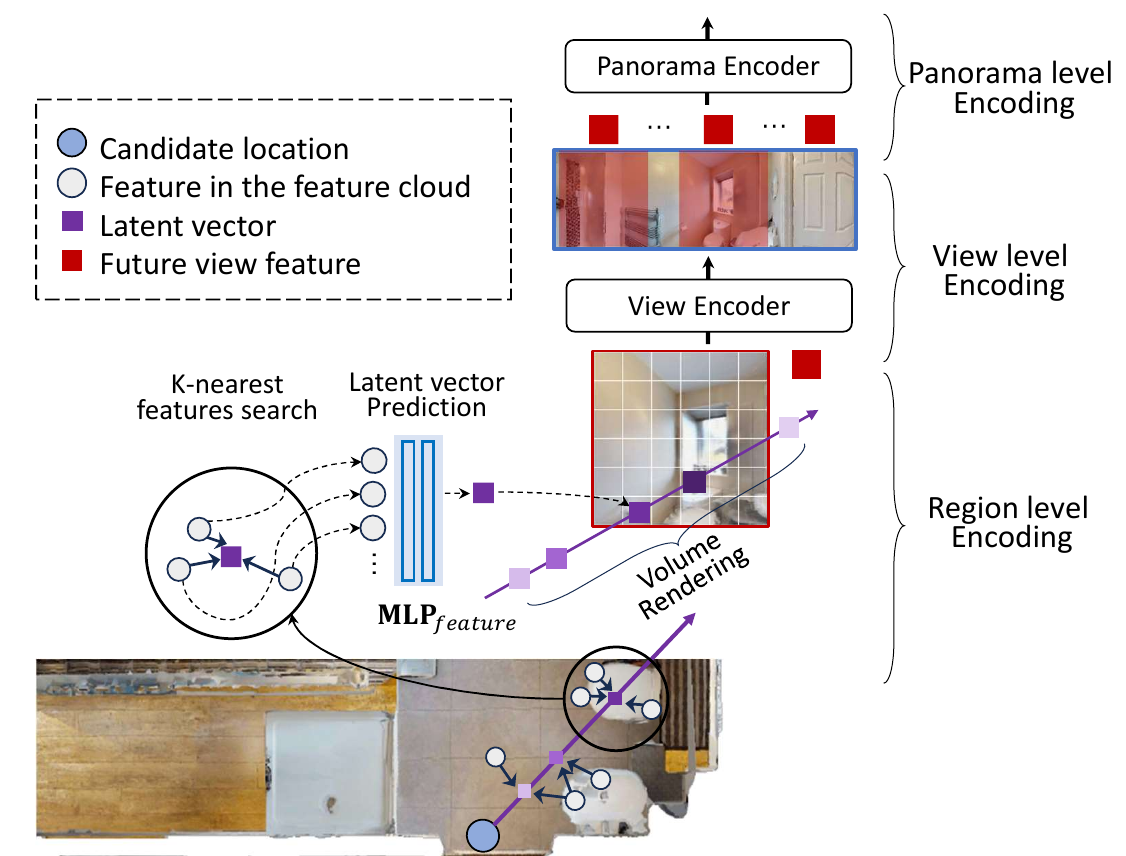}
\par\end{center}%
\end{minipage}
\vspace{-15pt}
\caption{Illustration of the volume rendering method and hierarchical encoding. 
}
\label{fig:nerf}
\vspace{-10pt}
\end{figure}

\subsubsection{Region Level Encoding via Volume Rendering}\label{sec:region_level_encoding}
As shown in Figure~\ref{fig:nerf}, to produce a feature map $\textbf{R} \in \mathbb{R}^{H \times W \times D}$ for each future view, given the encoded feature cloud $\mathcal{M}$ and the camera pose of the future view,  the HNR model predicts each region feature $\textbf{R}_{h,w}$ via volume rendering method~\cite{mildenhall2021nerf}. Specifically, for each region-level representation, the HNR model uniformly samples $N$ points $\{\mathcal{P}_n | n = 1, ..., N\}$ along the ray from the camera position $\mathcal{P}_1$ of the candidate location to the predicted region's center until 10 meters away. The KD-Tree algorithm~\cite{grandits_geasi_2021} is used to search the k-nearest features $\{\textbf{g}_{k}\}_{k=1}^K$ in feature cloud for each sampled point $\mathcal{P}_n=[\mathcal{X}_n,\mathcal{Y}_n,\mathcal{Z}_n]$. Then a $\textbf{MLP}_{feature}$ network is used to aggregate the k-nearest features of $\mathcal{P}_n$ within radius $R$ to produce a latent vector $\textbf{r}_n \in \mathbb{R}^{D}$ and the volume density $\sigma_n  \in \mathbb{R}^{1}$ as Equation~\ref{formula:latent_vector}.
These latent vectors along the ray are composited into the region feature $\textbf{R}_{h,w}$ via volume rendering as Equation~\ref{equ:region_feature}. 

To give the $\textbf{MLP}_{feature}$ network translational invariance for better generalization, the relative position $P^{rel}_{k}$ and relative orientation $\theta^{rel}_{k}$ of k-nearest feature $\textbf{g}_{k}$ to sampled point $\mathcal{P}_n$ is calculated as:

\vspace{-10pt}
\begin{align}
&P_k^{rel} = [x_{k}^{rel}\ ,\ y_{k}^{rel},\ z_{k}^{rel}]\notag\\&=[\ ({x}_{k}-\mathcal{X}_{n})\cdot cos\Theta_{n}+({y}_{k}-\mathcal{Y}_{n})\cdot sin\Theta_{n}\ ,\ \notag\\&\ \ \ \ \ \ \ \ ({y}_{k}-\mathcal{Y}_{n})\cdot cos\Theta_{n}-({x}_{k}-\mathcal{X}_{n})\cdot sin\Theta_{n}\ ,\ \notag\\&\ \ \ \ \ \ \ \ \ 
z_{k}-\mathcal{Z}_n\ ]\label{formula:relative_coordinates} 
\end{align}

\vspace{-20pt}

\begin{align}
&\theta^{rel}_{k} = [\ \sin(\theta_{k}-\Theta_n),\ \cos(\theta_{k}-\Theta_n)\ ]\label{formula:relative_orientation}
\end{align}

\vspace{-5pt}
\noindent where $P_{k}=[x_k,y_k,z_k]$ is the world position of k-nearest feature $\textbf{g}_{k}$, $\Theta_n$ denotes the horizontal orientation of the rendered ray. $\theta_{k}$ and $s_{k}$ are the orientation and region size of $\textbf{g}_{k}$. The positional embedding is encoded as:

\vspace{-10pt}
\begin{align}
&\textbf{q}_k=\textbf{LN}(\textbf{W}_{1}[P^{rel}_{k},\theta^{rel}_{k},s_{k}])
\label{formula:positional_embedding}
\end{align}

\vspace{-6pt}

\noindent where $\textbf{LN}$ denotes layer normalization and $\textbf{W}_{1}$ is learnable parameters.
With k-nearest features $\{\textbf{g}_{k}\}_{k=1}^K$ and positional embeddings $\{\textbf{q}_{k}\}_{k=1}^K$, the $\textbf{MLP}_{feature}$ network aggregate all k-nearest features of sampled point $\mathcal{P}_n$ as follows:

\vspace{-12pt}
\begin{align}
&[\textbf{r}_n,\sigma_n] = \textbf{MLP}_{feature}(\{\textbf{g}_{k}\oplus\textbf{q}_k\}_{k=1}^K)\label{formula:latent_vector}
\end{align}

\vspace{-5pt}
The volume density $\sigma_n$~\cite{mildenhall2021nerf} can be interpreted as the differential probability of a ray terminating at point $\mathcal{P}_n$. As shown in Figure~\ref{fig:nerf}, the latent vector with a higher volume density has a higher contribution to the region feature. To reduce the computational cost, we adopt a \textit{sparse sampling} strategy. Specifically, if the sampled point $\mathcal{P}_n$ does not have any neighboring features within the radius $\hat{R}$, the latent vector $\textbf{r}_n$ and volume density $\sigma_n$ will be directly set as zero, without using $\textbf{MLP}_{feature}$. 

With all latent vectors $\{\textbf{r}\}_{n=1}^N$ of $N$ sampled points, we use the volume rendering method~\cite{mildenhall2021nerf} to produce a region feature $\textbf{R}_{h,w}$ for future view as follows:

\vspace{-12pt}
\begin{align}
\textbf{R}_{h,w} = \sum_{n=1}^{N}\tau_n(1-\exp(-\sigma_n\Delta_n))\textbf{r}_n, \notag \\ \text{where}\ \tau_n = \exp({-\sum_{i=1}^{n-1}\sigma_i\Delta_i})\label{equ:region_feature}
\end{align}

\vspace{-8pt}
\noindent $\tau_{n}$ represents volume transmittance, and $\Delta_n$ is the distance between adjacent sampled points. To enhance interaction among different region features, we use both region-level semantic alignment $\mathcal{L}_{region}$ and view-level semantic alignment $\mathcal{L}_{view}$ after view level encoding in Section~\ref{sec:view_level_encoding}.

Although the goal of region-level encoding is to generate regional semantic features, for image reconstruction and depth estimation, we also trained an $\textbf{MLP}_{rgbd}$ network to predict the color value $\textbf{c}_{n}$ and volume density $\hat{\sigma}_{n}$, then obtain the volume transmittance $\hat{\tau}$. As shown in Equation~\ref{equ:get_rgb} and \ref{equ:get_depth}, the RGB pixel $\textbf{C}_{h,w}$ and depth value $\textbf{D}_{h,w}$ can be predicted via volume rendering. The rendering loss $\mathcal{L}_{rgbd}$ is the squared error between rendered pixels and ground truth. 

\vspace{-12pt}
\begin{align}
\textbf{C}_{h,w} = \sum_{n=1}^{N}\hat{\tau}_n(1-\exp(-\hat{\sigma}_n\Delta_n))\textbf{c}_n\label{equ:get_rgb}
\end{align}
\vspace{-20pt}
\begin{align}
\textbf{D}_{h,w} = \sum_{n=1}^{N}\hat{\tau}_n(1-\exp(-\hat{\sigma}_n\Delta_n))\textbf{d}_n\label{equ:get_depth}
\end{align} 

\vspace{-8pt}
\noindent where $\textbf{c}_n$ denotes color values, and $\textbf{d}_n$ denotes the distance between the sampled point $\mathcal{P}_n$ and the camera position $\mathcal{P}_1$.

\subsubsection{View Level Encoding}\label{sec:view_level_encoding}
The regional feature $\textbf{R}_{h,w}$ obtained in Section~\ref{sec:region_level_encoding} can only represent a small-scale region. To represent the entire future view and predict features of empty regions by integrating surrounding contexts, the regional feature map $\textbf{R}$ together with a learnable view token $\textbf{V}$ is inputted into the view encoder and output the encoded $\hat{\textbf{\textbf{R}}}$ and $\hat{\textbf{V}}$. The view encoder consists of four-layer transformers.

As shown in Figure~\ref{fig:framework}, to supervise the encoded region features $\hat{\textbf{R}}$, we align each region feature $\hat{\textbf{R}}_{h,w}$ with the CLIP embedding $\textbf{R}^{gt}_{h,w}$ extracted from a small-scale cropped image within the actual future view. Similarly, the view feature $\hat{\textbf{V}}$ is aligned with the CLIP embedding $\textbf{V}^{gt}$ extracted from an entire future view. During training, we randomly sample some region features and then minimize the loss between predicted features and actual CLIP embeddings, by maximizing cosine similarity as follows:

\vspace{-12pt}
\begin{align}
\mathcal{L}_{region} &= \sum_{h,w}(1-\frac{\hat{\textbf{R}}_{h,w}\cdot \textbf{R}^{gt}_{h,w}}{\Vert \hat{\textbf{R}}_{h,w}\Vert \Vert \textbf{R}^{gt}_{h,w} \Vert}) \\
\mathcal{L}_{view} &= 1-\frac{\hat{\textbf{V}}\cdot \textbf{V}^{gt}}{\Vert \hat{\textbf{V}}\Vert \Vert \textbf{V}^{gt} \Vert}
\end{align}

\vspace{-5pt}
For better generalization ability, the HNR model is pre-trained in large-scale HM3D~\cite{ramakrishnan2021habitat} dataset with 800 training scenes. Specifically, we randomly select a starting location in the scene and randomly move to a navigable candidate location at each step. At each step, parts of the unvisited candidate locations are randomly picked to predict future views. The model is optimized with the training loss $\alpha\mathcal{L}_{rgbd}+\beta\mathcal{L}_{region}+\gamma\mathcal{L}_{view}$, where $\alpha,\beta,\gamma$ are the factors of proportionality. The maximum number of action steps per episode is set to 15.

\subsubsection{Panorama Level Encoding}
The pre-trained HNR model in Section~\ref{sec:view_level_encoding} can be used to predict future view representations and combine them as panoramic representations of candidate locations through the panorama encoder, which is  described in Figure~\ref{fig:nerf}. Given a navigable candidate location, the HNR model predicts 12 single-view semantic features and depth maps at 30 degrees separation. Furthermore, the waypoint predictor model~\cite{Hong2022bridging} is used to predict navigable locations around the candidate location via these depth maps. The predicted future view features and navigable locations are encoded by the panorama encoder for lookahead exploration.


\subsection{Architecture of the Lookahead VLN model}\label{sec:lookahead_vln_model}
\begin{figure}
\noindent\begin{minipage}[b]{1\columnwidth}%
\begin{center}
\includegraphics[width=1.\columnwidth]{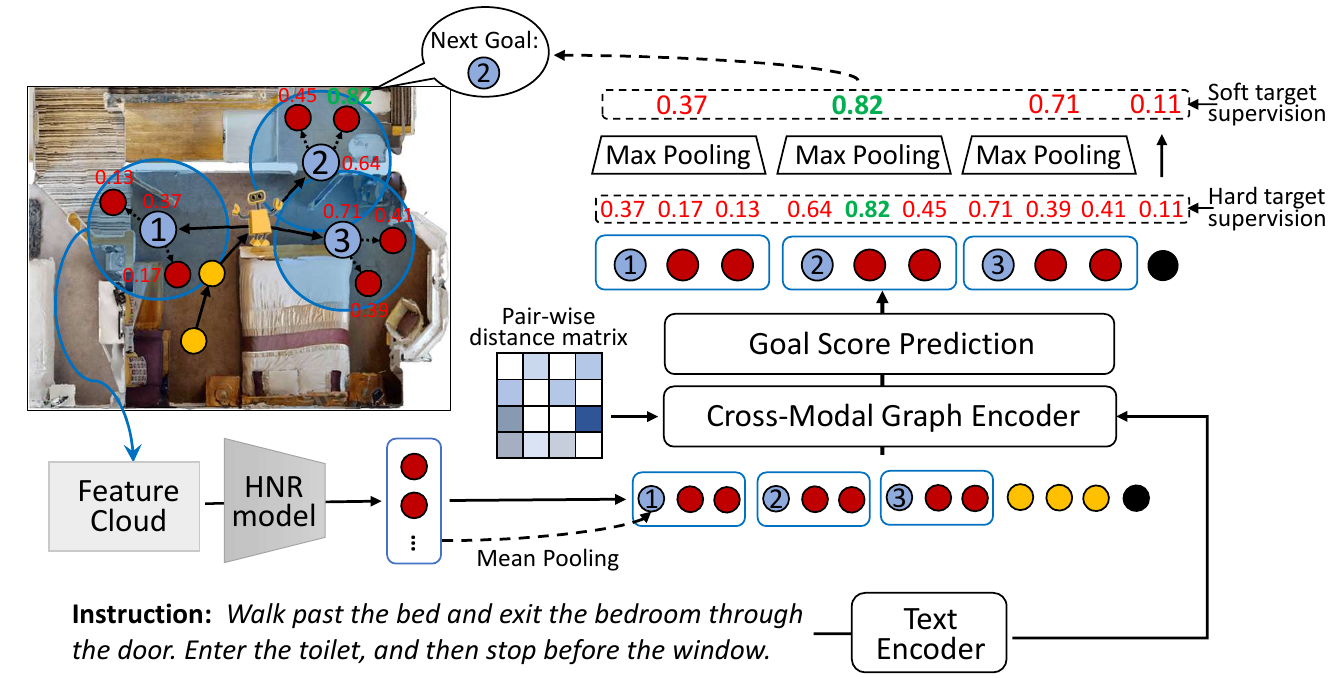}
\par\end{center}%
\end{minipage}
\vspace{-15pt}
\caption{The framework of the lookahead VLN model. In addition to the stop embedding (black), three types of nodes are used to structure the topological map: visited nodes (yellow), candidate nodes (blue) and lookahead nodes (red).}
\label{fig:vln}
\vspace{-10pt}
\end{figure}

\subsubsection{Node Embedding}\label{sec:node_embedding}
Three types of nodes are utilized to structure the topological map, as shown in Figure~\ref{fig:vln}. These nodes include visited nodes (yellow), candidate nodes (blue), and lookahead nodes (red). 
$\mathcal{V}^{visited}_t$ is represented as the current node embedding via average pooling 12 observed view representations. $\{\mathcal{V}^{visited}_{i}\}_{i=1}^t$ are all the visited node embeddings.

For each navigable candidate location, the HNR model predicts 12 future view features $\hat{\textbf{V}}$ at 30 degrees separation. 
The panorama encoder consisting of a two-layer transformer is used to encode $\hat{\textbf{V}}$ with corresponding position embedding and output features $\widetilde{\textbf{V}}$. $\mathcal{V}^{candidate}$ is represented as the candidate node embedding via average pooling of $\widetilde{\textbf{V}}$. The lookahead node embeddings are denoted as $\mathcal{V}^{lookahead}$, which are navigable parts of the future views $\widetilde{\textbf{V}}$ with corresponding position embedding. The lookahead nodes  connected to $\mathcal{V}^{candidate}_m$ are denoted as $\mathcal{V}^{lookahead}_{\mathcal{N}(m)}$.
The ‘stop’ embedding $\mathcal{V}^{stop}$ is added to the graph to denote a stop action and connect it with all other node embeddings.






\subsubsection{Cross-Modal Graph Encoding}\label{sec:cross-modal_graph_encoding}
To encode the environmental topological map and evaluate future path branches in it, all node representations in Section~\ref{sec:node_embedding} are fed into a 4-layer cross-modal transformer to conduct interaction. Each transformer layer consists of a cross-attention layer and a graph-aware self-attention layer (GASA). For cross-attention calculation, the encoded visited nodes $\mathcal{V}^{visited}$, candidate nodes $\mathcal{V}^{candidate}$, lookahead nodes $\mathcal{V}^{lookahead}$ and stop embedding $\mathcal{V}^{stop}$ are used as query tokens, the word embeddings $\mathcal{W}$ are used as key and value tokens.
The standard self-attention layer only considers visual similarity among nodes, which may overlook nearby nodes that are more relevant than distant nodes. To this end, following~\cite{chen2022think-GL,an2023etpnav}, we adopt the GASA layer that further takes into account the graph topology when computing inter-node attention for node encoding:

\vspace{-8pt}
\begin{align}
\textrm{GASA}(\mathcal{V})=\textrm{Softmax}(\frac{\mathcal{V} \textbf{W}_\text{q} (\mathcal{V} \textbf{W}_\text{k})^T}{\sqrt{d}}+E\textbf{W}_\text{e})\mathcal{V} \textbf{W}_\text{v}
\end{align}

\vspace{-5pt}
\noindent where $\mathcal{V}$ denotes node representations, $E$ is the pair-wise distance matrix obtained from the topological map, and $\textbf{W}_\text{q},\textbf{W}_\text{k},\textbf{W}_\text{e},\textbf{W}_\text{v}$ are learnable parameters. After cross-modal graph encoding, the encoded node representations are denoted as $\hat{\mathcal{V}}^{visited}$, $\hat{\mathcal{V}}^{candidate}$, $\hat{\mathcal{V}}^{lookahead}$, and $\hat{\mathcal{V}}^{stop}$.

\subsubsection{Lookahead Exploration and Action Prediction}\label{sec:lookahead}
The model predicts a navigation goal score for each node in the topological map as follows:

\begin{align}
S = \textbf{FFN} (\hat{\mathcal{V}})
\end{align}

\noindent where \textbf{FFN} denotes a feed-forward network. To avoid unnecessary repeated visits to visited nodes, the scores for visited nodes are masked. With the candidate node $\hat{\mathcal{V}}^{candidate}_m$ and those lookahead nodes $\hat{\mathcal{V}}^{lookahead}_{\mathcal{N}(m)}$ connected to it, the VLN model predicts their goal scores $[S^{candidate}_m, S^{lookahead}_{\mathcal{N}(m)}]$ and take the maximum value among them as the goal score of the $m$-th path branch:

\vspace{-12pt}
\begin{align}
S^{path} = \{\textrm{Max} ([S^{candidate}_m,S^{lookahead}_{\mathcal{N}(m)}])\}_m^M\label{path_scores}
\end{align}

\vspace{-5pt}
Finally, the agent selects a candidate node to move according to the predicted goal scores (\textit{i.e.}, select the path branch with the maximum score). The agent performs the stop action if the `stop' score has the maximum value.
If the selected candidate node is not adjacent to the current node, the agent computes the shortest path to the goal by performing Dikjstra's algorithm on the topological map. The control module~\cite{an2023etpnav} is responsible for converting the topological plan into a series of low-level actions that guide the agent to the selected candidate node.


As shown in Figure~\ref{fig:vln}, there are two navigation supervision strategies. The first one is the \textit{Hard target} supervision $\mathcal{A}_{hard}$, which selects an unvisited node (candidate node or lookahead node) with the shortest distance to the destination. Then goal scores of all unvisited nodes are used to calculate the cross-entropy loss. The other one is the \textit{Soft target} supervision $\mathcal{A}_{soft}$, which selects a candidate node with the shortest distance to the destination. The goal scores $S^{path}$ (in Equation~\ref{path_scores}) of candidate nodes are calculated by max pooling all scores of the corresponding path branch shown in Figure~\ref{fig:vln}. Then goal scores after max pooling are used to calculate the cross-entropy loss:

\vspace{-12pt}
\begin{align}
\mathcal{L}_{nav}=\textrm{CrossEntropy}(S^{path},\mathcal{A}_{soft})
\end{align}

\vspace{-8pt}

%% file: sec/4_experiment.tex
\section{Experiment}
\label{sec:experiment}

\begin{table*}
\small
\tabcolsep=0.1cm
\centering{}%
\begin{tabular}{c|ccccc|ccccc|ccccc}
\hline 
\multirow{2}{*}{Methods} & \multicolumn{5}{c|}{Val Seen} & \multicolumn{5}{c|}{Val Unseen} & \multicolumn{5}{c}{Test Unseen}\tabularnewline
\cline{2-16} \cline{3-16} \cline{4-16} \cline{5-16} \cline{6-16} \cline{7-16} \cline{8-16} \cline{9-16} \cline{10-16} \cline{11-16} \cline{12-16} \cline{13-16} \cline{14-16} \cline{15-16} \cline{16-16}
 & TL\textdownarrow{} & NE\textdownarrow{} & OSR\textuparrow{} & SR\textuparrow{} & SPL\textuparrow{} & TL\textdownarrow{} & NE\textdownarrow{} & OSR\textuparrow{} & SR\textuparrow{} & SPL\textuparrow{} & TL\textdownarrow{} & NE\textdownarrow{} & OSR\textuparrow{} & SR\textuparrow{} & SPL\textuparrow{}\tabularnewline
\hline

 




CM$^{2}$\ \cite{georgakis2022cm2} & 12.05 & 6.10 & 50.7 & 42.9 & 34.8 & 11.54 & 7.02 & 41.5 & 34.3 & 27.6  & 13.9 & 7.7 & 39 & 31 & 24\tabularnewline


WS-MGMap\ \cite{chen2022weakly} & 10.12 & 5.65 & 51.7 & 46.9 & 43.4 & 10.00 & 6.28 & 47.6 & 38.9 & 34.3 & 12.30 & 7.11 & 45 & 35 & 28\tabularnewline

Sim-2-Sim\ \cite{krantz2022sim2sim} & 11.18 & 4.67 & 61 & 52 & 44 & 10.69 & 6.07 & 52 & 43 & 36 & 11.43 & 6.17 & 52 & 44 & 37\tabularnewline



CWP-RecBERT\textcolor{black}\ \cite{Hong2022bridging} & 12.50 & 5.02 & 59 & 50 & 44 & 12.23 & 5.74 & 53 & 44 & 39 & 13.31 & 5.89 & 51 & 42 & 36\tabularnewline 

GridMM\textcolor{black}\
\cite{wang2023gridmm}& 12.69 & 4.21 & 69 & 59 & 51 & 13.36 & 5.11 & 61 & 49 & 41 & 13.31 & 5.64 & 56 & 46 & 39
\tabularnewline
 
Reborn\textcolor{black}\ \cite{an20221st} &  10.29 &  4.34 & 67 & 59 & 56 & 10.06 & 5.40 &  57 & 50 & 46 & 11.47 & 5.55 & 57 & 49 & 45\tabularnewline

Ego$^{2}$-Map\textcolor{black}\ \cite{hong2023learning} & - & - & - & - & - & - & 4.94 & - & 52 & 46 & 13.05 & 5.54 & 56 & 47 & 41 \tabularnewline

DREAMWALKER\textcolor{black}\ \cite{wang2023dreamwalker} & 11.6 & 4.09 & 66 & 59 & 48 & 11.3 & 5.53 & 59 & 49 & 44 & 11.8 & 5.48 & 57 & 49 & 44 \tabularnewline

ScaleVLN\textcolor{black}\ \cite{wang2023scaling} & - & - & - & - & - & - & 4.80 & - & 55 & \textbf{51} & - & 5.11 & - & 55 & \textbf{50} \tabularnewline

BEVBert\textcolor{black}\ \cite{an2023bevbert} & - & - & - & - & - & - & 4.57 & \textbf{67} & 59 & 50 & - & \textbf{4.70} & \textbf{67} & \textbf{59} & \textbf{50}\tabularnewline

ETPNav\textcolor{black}\ \cite{an2023etpnav} &  11.78 & 3.95 & 72 & 66 & 59 & 11.99 & 4.71 & 65 & 57 & 49 & 12.87 & 5.12 & 63 & 55 & 48\tabularnewline

\hline

\textcolor{black} HNR (Ours) & 11.79 & \textbf{3.67} & \textbf{76} & \textbf{69} & \textbf{61} & 12.64 & \textbf{4.42} & \textbf{67} & \textbf{61} & \textbf{51} & 13.03 & 4.81 & \textbf{67} & 58 & \textbf{50}
 \tabularnewline
\hline 
\end{tabular}
\vspace{-8pt}
\caption{Evaluation on the R2R-CE dataset.}\label{R2R-CE_sota}
\end{table*}

\begin{table*}
\small
\tabcolsep=0.07cm
\centering{}%
\begin{tabular}{c|ccccc|ccccc|ccccc}
\hline 
\multirow{2}{*}{Methods} & \multicolumn{5}{c|}{Val Seen} & \multicolumn{5}{c|}{Val Unseen} & \multicolumn{5}{c}{Test Unseen}\tabularnewline
\cline{2-16} \cline{3-16} \cline{4-16} \cline{5-16} \cline{6-16} \cline{7-16} \cline{8-16} \cline{9-16} \cline{10-16} \cline{11-16} \cline{12-16} \cline{13-16} \cline{14-16} \cline{15-16} \cline{16-16}
 & NE\textdownarrow{} & SR\textuparrow{} & SPL\textuparrow{} & NDTW\textuparrow{} & SDTW\textuparrow{} & NE\textdownarrow{} & SR\textuparrow{} & SPL\textuparrow{} & NDTW\textuparrow{} & SDTW\textuparrow{} & NE\textdownarrow{} & SR\textuparrow{} & SPL\textuparrow{} & NDTW\textuparrow{} & SDTW\textuparrow{}\tabularnewline
\hline 

CWP-CMA\textcolor{black}~\cite{Hong2022bridging} & - & - & - & - & - &  8.76 & 26.59 & 22.16 & 47.05 & - & 10.40 & 24.08 & 19.07 & 37.39 & 18.65\tabularnewline
\hline

CWP-RecBERT\textcolor{black}~\cite{Hong2022bridging} & - & - & - & - & - & 8.98 & 27.08 & 22.65 & 46.71 & - & 10.40 & 24.85 & 19.61 & 37.30 & 19.05\tabularnewline
\hline

Reborn\textcolor{black}~\cite{an20221st} & 5.69 & 52.43 & 45.46 & 66.27 & 44.47 & 5.98 & 48.60 & 42.05 & 63.35 & 41.82 & 7.10 & 45.82 & 38.82 & 55.43 & 38.42\tabularnewline
\hline

ETPNav\textcolor{black}~\cite{an2023etpnav} & 5.03 & 61.46 & 50.83 & 66.41 & 51.28 & 5.64 & 54.79 & 44.89 & 61.90 & 45.33 & 6.99 & 51.21 & 39.86 & 54.11 & 41.30\tabularnewline
\hline

\textcolor{black} HNR (Ours) & \textbf{4.85} & \textbf{63.72} & \textbf{53.17} & \textbf{68.81} & \textbf{52.78} & \textbf{5.51} & \textbf{56.39} & \textbf{46.73} & \textbf{63.56} & \textbf{47.24} & \textbf{6.81} & \textbf{53.22} & \textbf{41.14} & \textbf{55.61} & \textbf{42.89}
 \tabularnewline
\hline 
\end{tabular}
\vspace{-8pt}
\caption{Evaluation on the RxR-CE dataset.}\label{RxR-CE_sota}
\vspace{-10pt}
\end{table*}

\subsection{Datasets and Evaluation Metrics}
We evaluate our model on the R2R-CE~\cite{Krantz2020r2r-ce} and RxR-CE~\cite{2020_RXR} datasets in continuous environments. 

\textbf{R2R-CE}~\cite{Krantz2020r2r-ce} 
is collected based on the
discrete Matterport3D environments~\cite{matterport3d} 
with the Habitat simulator~\cite{ramakrishnan2021habitat}, enabling the agent to navigate in the continuous environments. It provides step-by-step instructions, and the average length is 32 words. The agent uses a $15^{\circ}$ turning angle and the horizontal field-of-view is $90^{\circ}$.

\textbf{RxR-CE}~\cite{2020_RXR} is a larger multilingual VLN dataset containing 126K instructions in English, Hindi, and Telugu. It includes trajectories that are diverse in terms of length (average is 15 meters), which is more challenging in the continuous environments. The agent uses a $30^{\circ}$ turning angle and the horizontal field-of-view is $79^{\circ}$.

There are several standard metrics~\cite{VLN_2018vision} in
VLN for evaluating the agent’s performance, including Trajectory Length (TL), Navigation Error (NE), Success Rate
(SR), SR given the Oracle stop policy (OSR), Normalized inverse of the Path Length (SPL), Normalized Dynamic Time Warping (nDTW), and Success weighted by normalized Dynamic Time Warping (SDTW).

\subsection{Comparison to State-of-the-Art Methods}
Table~\ref{R2R-CE_sota} and \ref{RxR-CE_sota} 
represent the performance of our proposed HNR model compared with existing VLN models on the R2R-CE and RxR-CE datasets respectively.
Overall, HNR achieves state-of-the-art results in the majority of metrics, demonstrating the effectiveness of the proposed approach from diverse perspectives. As illustrated in Table~\ref{R2R-CE_sota}, for the val unseen split  of the R2R-CE dataset, our model outperforms our baseline method ETPNav~\cite{chen2022think-GL} by 4\% on SR and 2\% on SPL. For the test unseen split, the proposed method outperforms ETPNav by 3\% on SR and 2\% on SPL. Meanwhile, as illustrated in Table~\ref{RxR-CE_sota}, the proposed method also achieves the improvement of 2\% in the majority of metrics on the RxR-CE dataset.

Compared with DREAMWALKER~\cite{wang2023dreamwalker} in Table~\ref{R2R-CE_sota}, which adopts a similar idea of lookahead exploration, our HNR model achieves performance improvement of about 10\% on SR for all splits. On the one hand, our HNR constructs a unified future path tree to represent the future environment, which is better than the step-by-step path search strategy used in DREAMWALKER. The step-by-step search (\textit{i.e.}, only evaluate one path branch at a time) makes it difficult to compare different future paths. On the other hand, the future environmental representations predicted by our HNR model are superior to those from the image generation model adopted by DREAMWALKER.




\subsection{Ablation Study}\label{sec:ablation_study}

As described in Section~\ref{sec:intro}, the representations of future environments play a crucial role in lookahead exploration. 
In this section, we compare several popular representation strategies for future environments. 
(1) The most usual single-view representations~\cite{an2023etpnav}, which rely on single-view visual observation of the current location to perceive candidates.  
(2) NeRF-based rendering methods~\cite{kwon2023renderable} that focus on rendering RGB pixels for future view images and extract 
features via the CLIP model. (3) Image Generation method~\cite{wang2023dreamwalker,koh2021pathdreamer} that imagine the panoramic RGBD images for candidate locations. (4) Our proposed hierarchical neural radiance representation model (HNR). 

\begin{table}[ht]
\small
\tabcolsep=0.08cm
\centering{}%
\begin{tabular}{cc|ccccc}
\hline 
\#& Representation Methods  & NE\textdownarrow{} & OSR\textuparrow{} & SR\textuparrow{} & SPL\textuparrow{}\tabularnewline
\hline

1& Single View  & 4.71 & 64.71 & 57.21 & 49.15\tabularnewline

2& NeRF Rendering  & 4.79 & 65.14 & 56.55 & 48.61\tabularnewline

3& Image Generation  & 4.68 & 66.01 & 58.35 & 50.96\tabularnewline

4& HNR  & 4.42 & 67.48 & 60.74 & 51.27\tabularnewline

5& Ground truth & 4.13 & 71.29 & 63.13 & 54.59\tabularnewline

6& HNR w/o $\mathcal{L}_{region}$ & 4.55 & 66.78 & 60.20 & 50.74\tabularnewline

7& HNR w/o Positional embeddings & 4.78 & 64.87 & 56.17 & 48.91\tabularnewline

\hline 
\end{tabular}
\vspace{-8pt}
\caption{Comparison among different candidate location representation methods on the val unseen split of the R2R-CE dataset.} \label{table:NeRF_comparison}
\vspace{-5pt}
\end{table}

\vspace{-10pt}
\paragraph*{
Comparisons among different representation methods.} 
For fair comparisons, 
rows 2-4 in Table~\ref{table:NeRF_comparison} adopt the same lookahead VLN model shown in Section~\ref{sec:lookahead_vln_model} for navigation planning but use different methods for future environments prediction. Traditional single-view representation (row 1) is restricted to a 
limited observable view and is difficult to comprehensively represent future environments due to the visual occlusions. The lookahead exploration strategy (rows 2-4) helps evaluate the agent’s future paths by anticipating the future environment for better navigation planning. NeRF-based RGB rendering method~\cite{kwon2023renderable} has a low image reconstruction accuracy in unseen environments due to the visual occlusions and high information redundancy of RGB images, resulting in undesirable performance gains. The image generation method~\cite{wang2023dreamwalker} has better performance gains than NeRF Rendering in unseen environments but is still inferior to our HNR model. To avoid the difficulty of pixel-level prediction, the proposed HNR model takes the CLIP embeddings as the prediction objective and predicts multi-level semantics through hierarchical encoding. 

\vspace{-10pt}
\paragraph*{
Upper bound of our lookahead exploration strategy.} The ground truth (GT) representations (row 5) are extracted from the actual future view images via the CLIP model. It demonstrates the upper bound of our lookahead VLN model with GT future view prediction and also confirms the effectiveness of the lookahead exploration strategy.

\vspace{-10pt}
\paragraph*{Region level semantic alignment.} 
As shown in Table~\ref{table:NeRF_comparison}, without the training objective $\mathcal{L}_{region}$ of region-level semantic alignment (row 6) in Section~\ref{sec:view_level_encoding}, the performance of the HNR model has degraded. Hierarchical encoding and multi-level semantic alignment help HNR integrate surrounding contexts and predict features of empty regions caused by visual occlusions.

\begin{figure}[ht]
\vspace{-5pt}
\noindent\begin{minipage}[ht]{1\columnwidth}%
\begin{center}
\includegraphics[width=0.8\columnwidth]{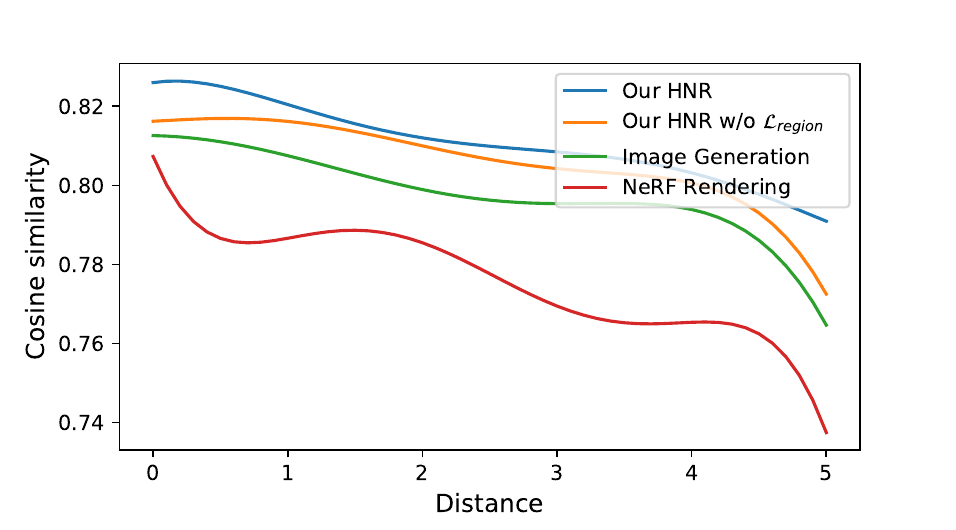}
\par\end{center}%
\end{minipage}
\vspace{-12pt}
\caption{Average cosine similarity between predicted future views and ground truth at different distances between candidate locations and agent, on the val unseen split of the R2R-CE dataset.}
\label{fig:5}
\vspace{-5pt}
\end{figure}

\vspace{-10pt}
\paragraph*{Quality analysis of future environment prediction.} As illustrated in Figure~\ref{fig:5}, we evaluate the quality of predicted representations by comparing the embeddings' cosine similarity between the predicted future views and ground truth. We observe a downward trend in cosine similarity as the distance between candidate location and agent increases. Combining Table~\ref{table:NeRF_comparison} and Figure~\ref{fig:5}, it's obvious that the better quality of future environmental representations makes the higher navigation performance gains. 
Using region-level alignment, our HNR method has the best representation quality and its cosine similarity is above 0.8 overall, supporting the lookahead exploration well within 5 meters.

\vspace{-10pt}
\paragraph*{Positional embeddings in region level encoding.} The positional embeddings calculated by Equation~\ref{formula:positional_embedding} represent the spatial relationships for region feature encoding. Without the position and orientation of the k-nearest features relative to the sampled point (row 7) in Table~\ref{table:NeRF_comparison}, $\textbf{MLP}_{feature}$ network cannot accurately estimate the volume density and fails to perceive spatial relations in the 3D environment, resulting in a significant performance decrease.

\vspace{-10pt}
\paragraph*{Number of k-nearest features.} Table~\ref{table:k_nearst} shows the effect of the number of k-nearest features on performance. We observe an upward trend in performance as the number of aggregated k-nearest features increases. Although the many more aggregated features make the region semantic prediction more accurate, the performance gains of the number over 4 are marginal, so we finally take the K value as 4 for lower computational cost.

\vspace{-10pt}
\paragraph*{Sparse sampling and k-nearest features search
.} Table~\ref{tab:time_2} illustrates two important strategies used in our HNR to improve computational efficiency. The first one is the sparse sampling strategy described in Section~\ref{sec:region_level_encoding}, the $\textbf{MLP}_{feature}$ is not used to aggregate features in empty areas of the feature cloud. Due to the high proportion of empty space in 3D indoor environments, this strategy can greatly reduce the redundant calculation. To parallelly search the k-nearest features of all sampled points, the KD-Tree algorithm is adopted in HNR, as described in Section~\ref{sec:region_level_encoding}. Since the k-nearest features search has a heavy computational cost in our HNR model, we use the CUDA implementation of KD-Tree algorithm~\cite{grandits_geasi_2021} to accelerate it, much faster than other nearest neighbor search methods~\cite{pedregosa2011scikit} (\textit{i.e.}, w/o KD-Tree) in Table~\ref{tab:time_2}.

\vspace{-10pt}
\paragraph*{Performance analysis of the lookahead exploration.}
In this part, we analyze different future path evaluation strategies and navigation supervision methods. As shown in Table~\ref{tab:lookahead_vln}, $S^{lookahead}$ indicates that the VLN model uses the scores of the lookahead nodes to evaluate the future paths, and $S^{candidate}$ indicates the usage of the candidate node scores, as shown in Equation~\ref{path_scores}. \textit{Soft target} and \textit{Hard target} are two different types of navigation supervision described in Section~\ref{sec:lookahead}. In Table~\ref{tab:lookahead_vln}, row 1 has the lowest performance without the lookahead exploration strategy. Row 2 doesn't use the lookahead node scores to evaluate the future paths and gain marginal performance improvement, confirming the necessity of the lookahead node representation. The performance of row 5 is superior to rows 3-4, showing the \textit{Soft target} supervision is the better navigation supervision. The lookahead node closest to the destination (\textit{i.e.}, \textit{Hard target}) is not sure of the highest semantic match score with the instruction due to visual occlusions and prediction noise, compulsively overfitting the \textit{Hard target} compromises the navigation performance.

\begin{table}[ht]
\small
\tabcolsep=0.08cm
\centering{}%
\begin{tabular}{c|ccccc}
\hline 
 Number of nearest features  & NE\textdownarrow{} & OSR\textuparrow{} & SR\textuparrow{} & SPL\textuparrow{}\tabularnewline
\hline

1 & 4.68 & 65.63 & 57.86 & 49.21\tabularnewline

2 & 4.57 & 66.99 & 59.49 & 50.81\tabularnewline

4 & 4.42 & 67.48 & 60.74 & 51.27\tabularnewline

6 & 4.37 & 67.70 & 60.58 & 51.30\tabularnewline

\hline 
\end{tabular}
\vspace{-8pt}
\caption{The effect of different numbers of nearest features in the HNR model on the val unseen split of the R2R-CE dataset.} \label{table:k_nearst}
\vspace{-5pt}
\end{table}

\begin{table}[ht]
\centering
\resizebox{1\linewidth}{!}{
\begin{tabular}{@{}cccc@{}}
\toprule
HNR & w/o Sparse sampling & w/o KD-Tree  \\ \midrule
87.3 Hz (11.5 ms)   & 5.9 Hz (169.5ms)   & 23.1 Hz (42.3 ms)     \\ \bottomrule
\end{tabular}}
\vspace{-8pt}
\caption{Runtime analysis measured on one RTX 3090 GPU.}
\label{tab:time_2}
\vspace{-5pt}
\end{table}

\begin{table}[ht]
\small
\noindent\begin{minipage}[t]{1\columnwidth}%
\tabcolsep=0.03cm
\begin{center}
\begin{tabular}{ccccc|ccccc}
\hline 
\# &$S^{lookahead}$ & $S^{candidate}$ & Soft target & Hard target & NE\textdownarrow{} & OSR\textuparrow{} & SR\textuparrow{} & SPL\textuparrow{}\tabularnewline
\hline 

1 & & & & & 4.71 & 64.71 & 57.21 & 49.15
\tabularnewline

2 & & $\checkmark$ & $\checkmark$ & & 4.71
 & 66.43 & 57.75 & 49.54
\tabularnewline

3 & $\checkmark$ & $\checkmark$ & & $\checkmark$ & 4.81 & 65.85 & 56.72
 & 48.41\tabularnewline
 
4 & $\checkmark$ & $\checkmark$ & $\checkmark$ &$\checkmark$& 4.62 & 67.92 & 59.87
 & 50.03\tabularnewline

\hline 
5 & $\checkmark$ & $\checkmark$ & $\checkmark$ & & 4.42 & 67.48 & 60.74 & 51.27\tabularnewline
\hline 
\end{tabular}
\par\end{center}%
\end{minipage}
\vspace{-8pt}
\caption{Ablation study of the lookahead VLN model.}\label{tab:lookahead_vln}
\vspace{-5pt}
\end{table}

%% file: sec/5_conclusion.tex
\section{Conclusion}
In this paper, we propose a lookahead exploration method for continuous vision-language navigation. Our proposed HNR model predicts multi-level future environmental representations through volume rendering and hierarchical encoding, supervised by CLIP embeddings. With predicted future representations, the lookahead VLN model constructs the navigable future path tree and selects the optimal path branch via efficient parallel evaluation. Extensive experiments demonstrate the accuracy of the HNR model's representations and the excellent performance of the lookahead VLN model. The lookahead exploration method helps navigation planning by predicting the future outcome of actions, which has great research potential for VLN and Embodied AI tasks.

\paragraph*{Acknowledgment.} This work was supported in part by
the National Natural Science Foundation of China under
Grants 62125207, 62102400, 62272436, and U23B2012, in part by the National Postdoctoral Program
for Innovative Talents under Grant BX20200338.

%% file: sec/X_suppl.tex
\clearpage

\appendix
\section*{Appendix}

\section{Experimental Details}
\label{sec:experimental_details}
\subsection{Settings of the HNR model}
\paragraph{Settings of volume rendering.} We set the k-nearest search radius $R$ as 1 meter, and the radius $\hat{R}$ for \textit{sparse sampling} strategy is also set as 1 meter. The rendered ray is uniformly sampled from 0 to 10 meters, and the number of sampled points $N$ is set as 256.

\begin{figure}
\noindent\begin{minipage}[ht]{1\columnwidth}%
\begin{center}
\includegraphics[width=0.4\columnwidth]{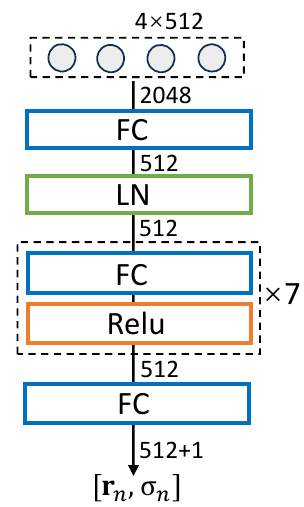}
\par\end{center}%
\end{minipage}
\vspace{-5pt}
\caption{Architecture of the $\textbf{MLP}_{feature}$ network. \textbf{FC} denotes a fully connected layer, \textbf{LN} denotes layer normalization and \textbf{Relu} denotes ReLU activation.}
\label{fig:sup_1}
\end{figure}

\paragraph{Settings of model architecture.} Since the number of k-nearest features is set to 4 and the dimension of each aggregated feature is 512, the input dimension of $\textbf{MLP}_{feature}$ network is 2048. The overall architecture of the $\textbf{MLP}_{feature}$ is shown in Figure~\ref{fig:sup_1}. The view encoder consists of four-layer transformers, and the number of region features within a future view is set as 7$\times$7.

\paragraph{Settings of pre-training.} 
The HNR model is pre-trained in large-scale HM3D~\cite{ramakrishnan2021habitat} dataset with 800 training scenes. Specifically, we randomly select a starting location in the scene and randomly move to a navigable candidate location at each step. At each step, up to 4 unvisited candidate locations are randomly picked to predict a future view in a random horizontal orientation, and 8 region features within it are randomly selected for region-level alignment. During pre-training, the horizontal field-of-view of each view is set as 90$^{\circ}$. The maximum number of action steps per episode is set to 15. Using 4 RTX3090 GPUs, the HNR model is pre-trained with a batch size of 4 and a learning rate 1e-4 for 20k episodes.

\subsection{Settings of the lookahead VLN model}
\paragraph{Settings of R2R-CE dataset.} The VLN model is initialized with the parameters of ETPNav~\cite{an2023etpnav} model trained in the R2R-CE dataset. Using 4 RTX3090 GPUs, the lookahead VLN model is trained with a batch size of 4 and a learning rate 1e-5 for 20k episodes.

\paragraph{Settings of RxR-CE dataset.} The VLN model is initialized with the parameters of ETPNav~\cite{an2023etpnav} model trained in the RxR-CE dataset. Using 4 RTX3090 GPUs, the lookahead VLN model is trained with a batch size of 4 and a learning rate 1e-5 for 100k episodes.

\section{Visualization and Examples}
\label{sec:visualization_and_examples}

\begin{figure*}[ht]
\makebox[\textwidth][c]
{\includegraphics[width=0.65\paperwidth]{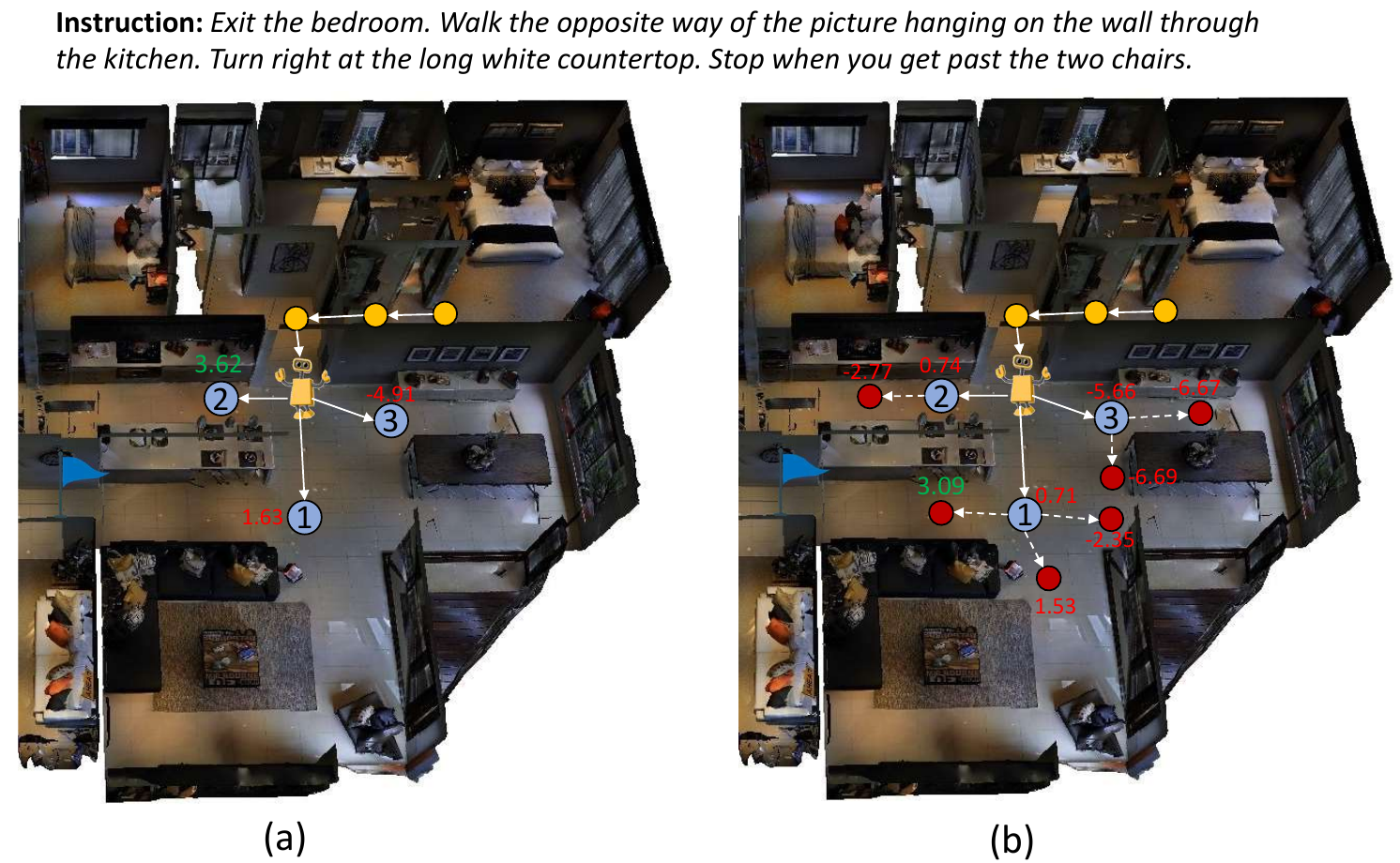}}
\vspace{-15pt}
\caption{A navigation example on the val unseen split of the R2R-CE dataset. (a) denotes the navigation strategy of ETPNav~\cite{an2023etpnav}. (b) denotes the lookahead exploration strategy of the lookahead VLN model.}
\label{fig:sup_2}
\end{figure*}

\begin{figure*}[ht]
\makebox[\textwidth][c]
{\includegraphics[width=0.65\paperwidth]{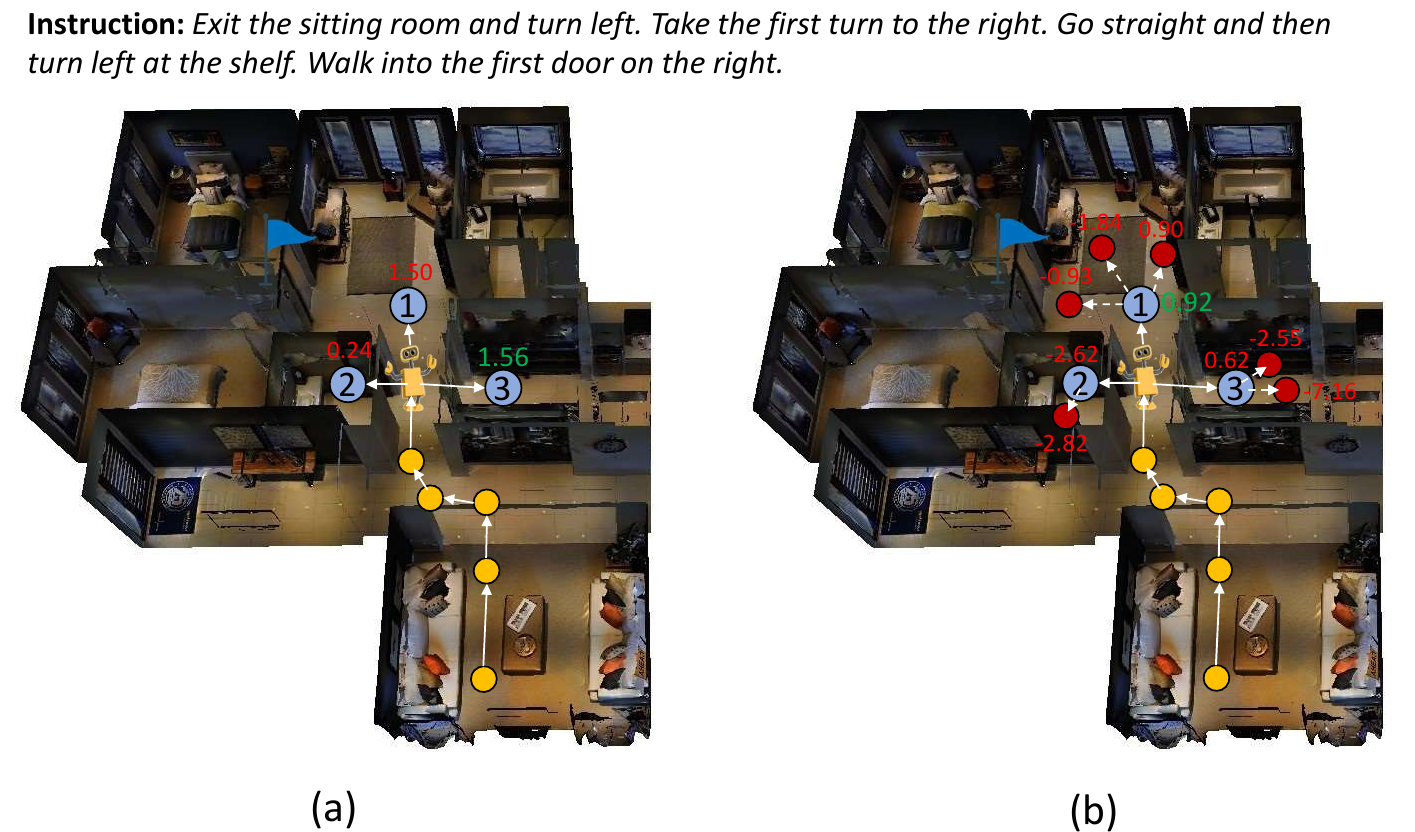}}
\vspace{-15pt}
\caption{A navigation example on the val unseen split of the R2R-CE dataset. (a) denotes the navigation strategy of ETPNav~\cite{an2023etpnav}. (b) denotes the lookahead exploration strategy of the lookahead VLN model.}
\label{fig:sup_3}
\end{figure*}

\paragraph{Visualization of the lookahead exploration strategy.}
Figure~\ref{fig:sup_2} and ~\ref{fig:sup_3} show examples that the HRN model with the lookahead exploration strategy has a more accurate evaluation for future paths than the ETPNav model.

\begin{figure*}
\makebox[\textwidth][c]
{\includegraphics[width=0.7\paperwidth]{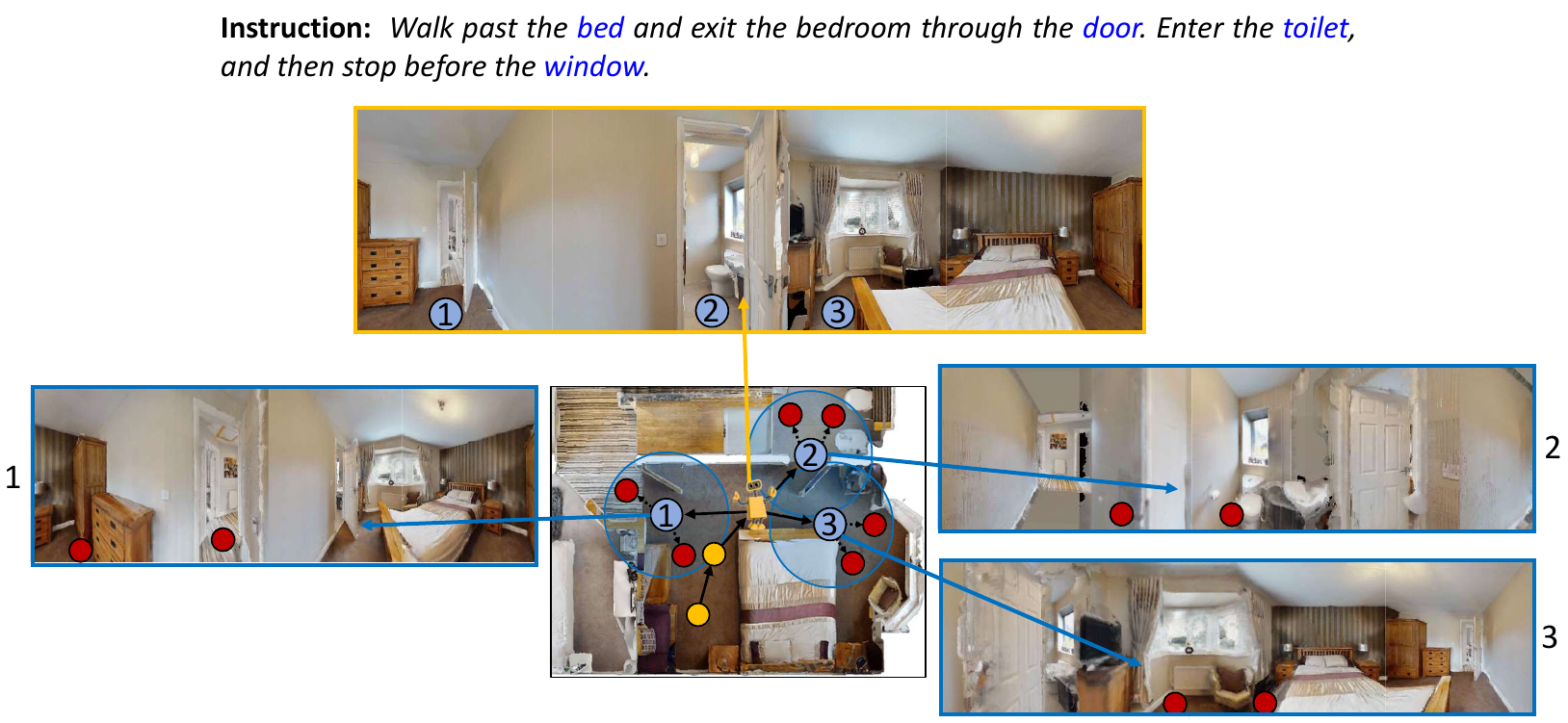}}
\vspace{-15pt}
\caption{Illustration of RGB reconstruction for candidate locations using the HNR model. The images in the yellow box are the agent's current observations. The images in the blue box are the rendered images for candidate locations using the HNR model.}
\label{fig:sup_4}
\end{figure*}

\paragraph{Visualization of the RGB reconstruction.}
Figure~\ref{fig:sup_4} illustrates the effect of RGB reconstruction for candidate locations using the HNR model. In fact, the grid features~\cite{wang2023gridmm} extracted by the CLIP model are not enough to enable the HNR model to render high-quality RGB images. Therefore, the additional point cloud is introduced, projected from the observed 224$\times$224 RGBD images during navigation, providing high-resolution geometry and texture-level details. The 4 nearest features and 16 nearest RGB points are fed into the $\textbf{MLP}_{rgbd}$ for RGB reconstruction and depth estimation. The $\textbf{MLP}_{feature}$ only takes 4 nearest features to predict the latent vector.

\begin{figure*}
\makebox[\textwidth][c]
{\includegraphics[width=0.7\paperwidth]{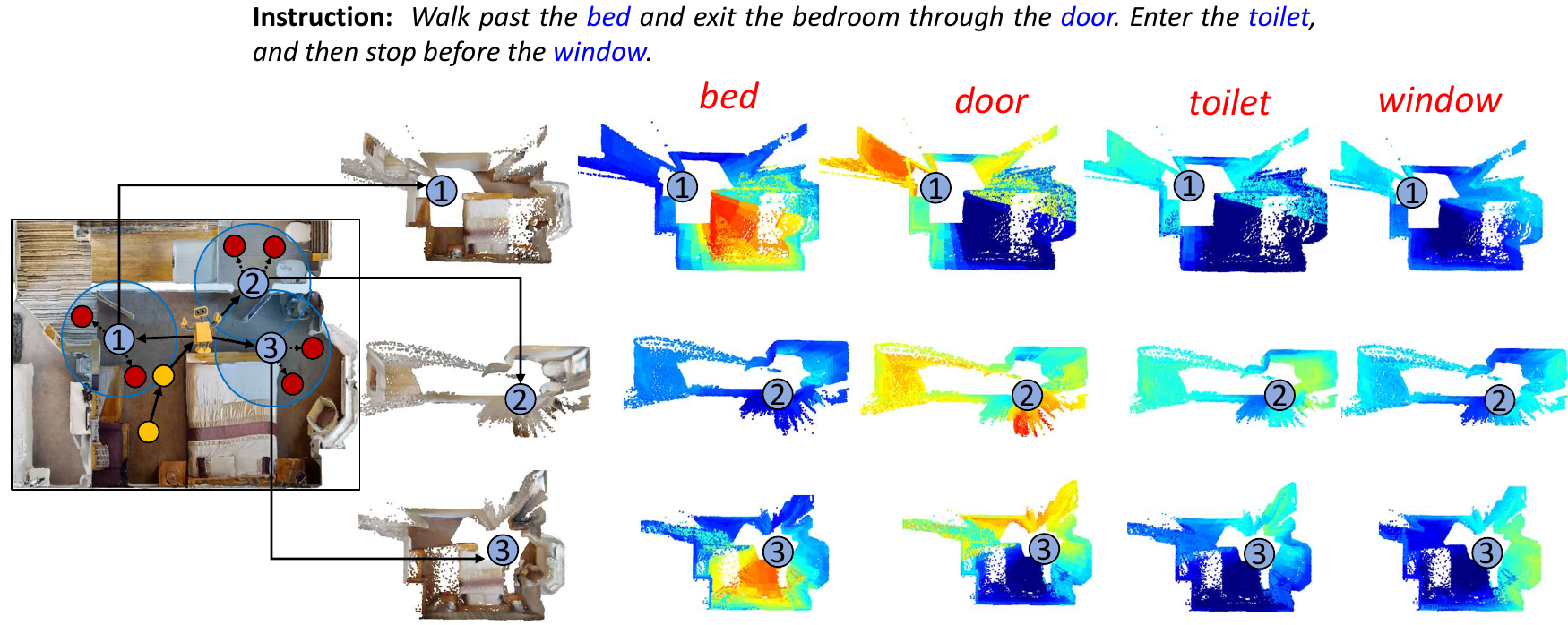}}
\vspace{-10pt}
\caption{Visualization of the predicted semantic features. The left part shows the top-down view of the reconstructed panoramas of the candidate locations. The right part shows the semantic similarity between the predicted region features and the specific language embeddings. The warmer color in the map represents a higher semantic similarity.}
\label{fig:sup_5}
\vspace{-5pt}
\end{figure*}

\begin{figure*}
\makebox[\textwidth][c]
{\includegraphics[width=0.7\paperwidth]{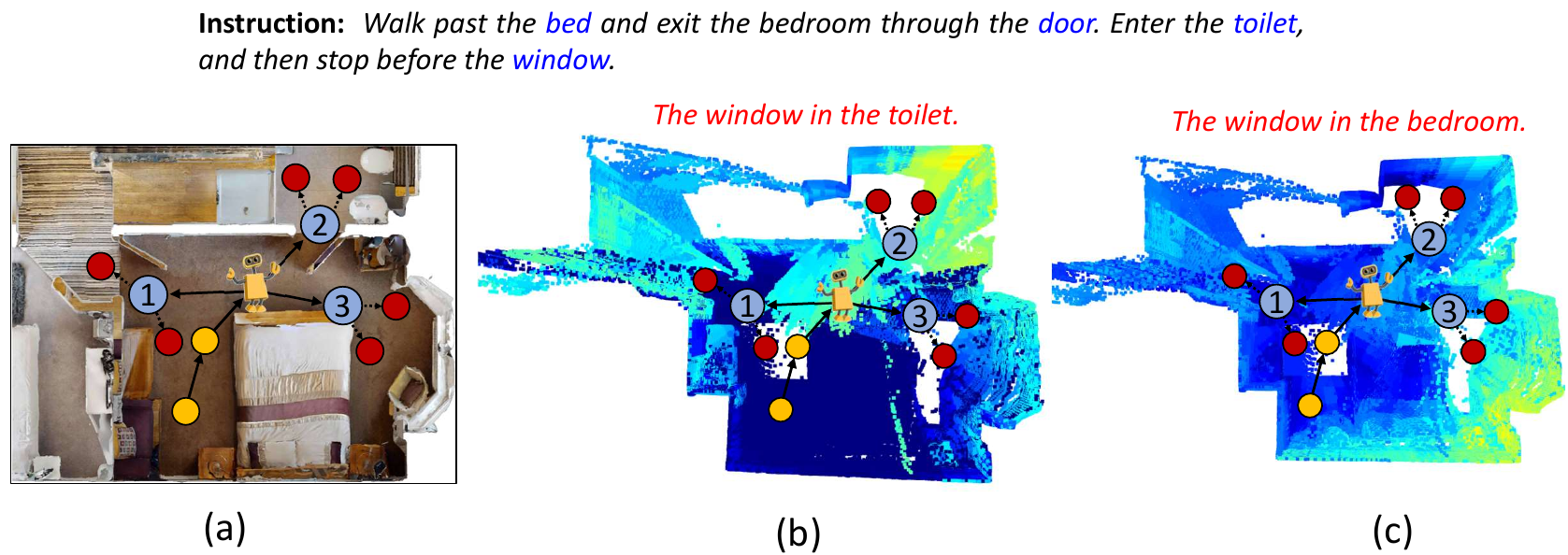}}
\vspace{-10pt}
\caption{Visualization of the predicted semantic features. (b) and (c) show the semantic similarity of region features to different sentences. The warmer color in the map represents a higher semantic similarity.}
\label{fig:sup_6}
\end{figure*}

\paragraph{Visualization of the predicted semantic features.} Figure~\ref{fig:sup_5} and~\ref{fig:sup_6} illustrate that the region features from the HNR model are well associated with the language by semantic alignment with the CLIP embeddings. As shown in Figure~\ref{fig:sup_5}, the predicted features surrounding the different candidate locations help the agent detect critical objects and understand the spatial relationships among them. In figure~\ref{fig:sup_6}, for semantic relationships between object and scene, the HNR model can also handle well. 

\begin{figure*}
\makebox[\textwidth][c]
{\includegraphics[width=0.65\paperwidth]{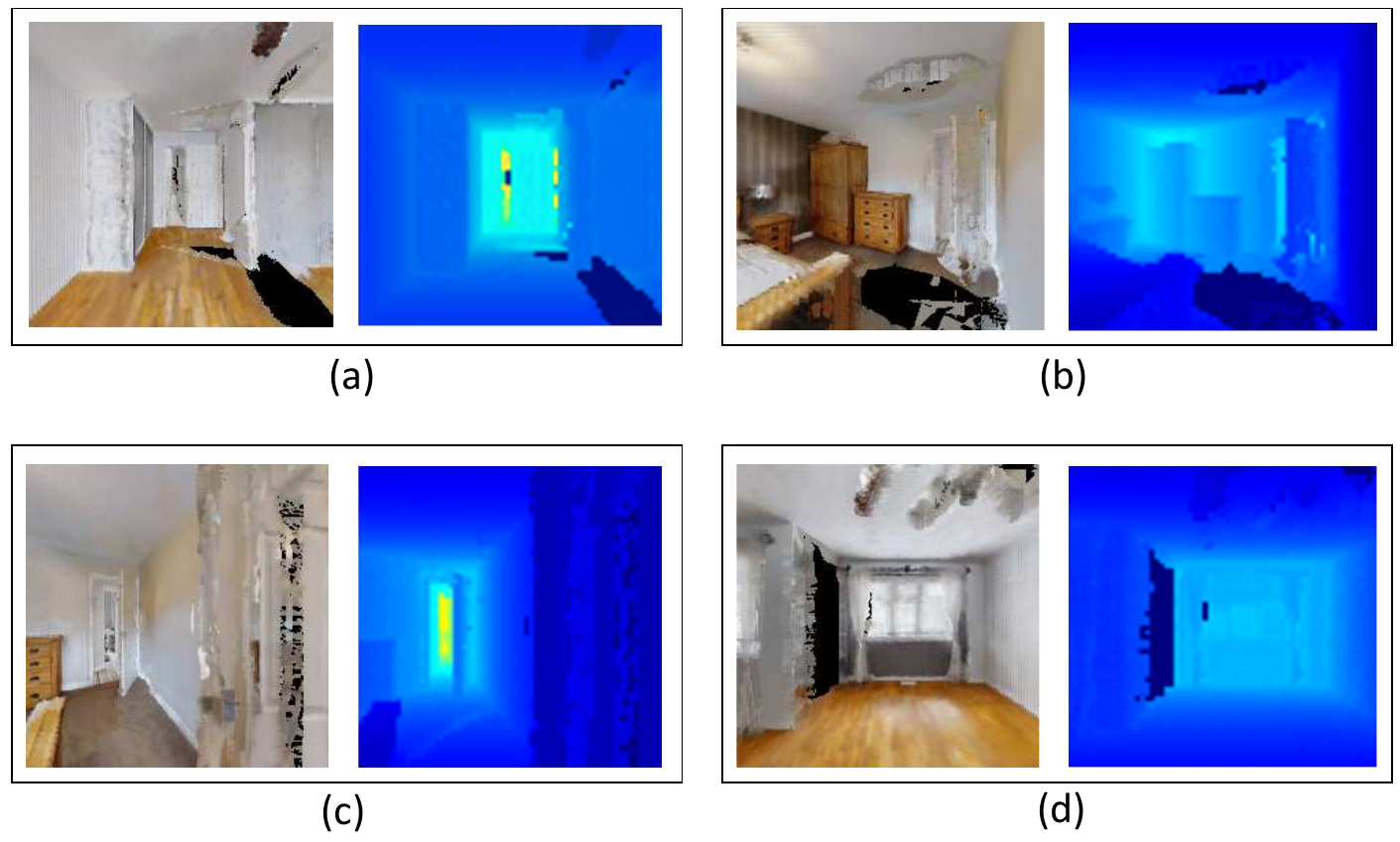}}
\vspace{-15pt}
\caption{Some failure cases of the rendered 224$\times$224 RGB images and 64$\times$64 depth images (please zoom in for better view).}
\label{fig:failure_cases}
\end{figure*}

\section{Some Discussions}

\paragraph*{Runtime for future view prediction.} 
Due to the large number of future path branches, the lookahead exploration requires extremely fast methods for predicting the future environment. Table~\ref{tab:time} shows the runtime for each future view prediction using different methods. The runtimes are measured on an NVIDIA GeForce RTX 3090 GPU. We can see that HNR achieves competitive inference speed and is fast enough for real-time lookahead exploration.

\begin{table}[ht]
\centering
\resizebox{1\linewidth}{!}{
\begin{tabular}{@{}cccc@{}}
\toprule
NeRF Rendering~\cite{kwon2023renderable} & Image Generation~\cite{wang2023dreamwalker} & HNR  \\ \midrule
21.6 Hz (46.3 ms)   & 12.6 Hz (79.4 ms)   & 87.3 Hz (11.5 ms)     \\ \bottomrule
\end{tabular}}
\vspace{-5pt}
\caption{Runtime analysis of different future view representation methods. NeRF Rendering and HNR generate a single view, while the Image Generation method generates an entire panorama.}
\label{tab:time}
\vspace{-5pt}
\end{table}

\paragraph{The input of lookahead exploration.} For each future view, HNR model predicts a 7$\times$7 region feature map using  $\textbf{MLP}_{feature}$ and a 64$\times$64 depth map using $\textbf{MLP}_{rgbd}$. Then the feature map is fed into the view encoder, and the depth map is upsampled
and fed into the waypoint predictor as described in Section 3.2.4. During navigation, the HNR model has not been used to reconstruct RGB images for lookahead exploration. The reasons are two-fold. \textbf{(1)} The computational cost of rendering 224$\times$224 RGB image exceeds that of predicting 7$\times$7 feature map by more than hundreds of times, which is unacceptable for real-time navigation. To further reduce the computational cost of training, in our experimental settings, the lookahead VLN model uses the depth map of ground truth for training and the rendered depth map for testing. \textbf{(2)} Due to visual occlusion during navigation, the RGB images reconstructed by the HNR model may still have empty regions and ghostly artifacts as shown in Figure~\ref{fig:failure_cases}, which introduce noisy visual features to the agent. The region feature map is more robust than the rendered RGB image, the hierarchical encoding and region-level semantic alignment are proposed to predict features of empty regions by integrating contexts in the view encoder. 

\paragraph{The limitations of the HNR model.} Although the future view features predicted by HNR work well for lookahead exploration, there is still room for improvement regarding the speed and quality of RGBD reconstruction. In the future, we will try faster 3D Gaussian Splatting~\cite{kerbl20233d}, and use the diffusion models~\cite{saharia2022palette} to fill in the empty region caused by visual occlusion. On the other hand, the 3D feature field~\cite{pmlr-v229-shen23a} with more geometric details is required for some Embodied AI tasks, such as mobile manipulation.

%% file: main.bbl
\begin{thebibliography}{51}
\providecommand{\natexlab}[1]{#1}
\providecommand{\url}[1]{\texttt{#1}}
\expandafter\ifx\csname urlstyle\endcsname\relax
  \providecommand{\doi}[1]{doi: #1}\else
  \providecommand{\doi}{doi: \begingroup \urlstyle{rm}\Url}\fi

\bibitem[Agarwal et~al.(2019)Agarwal, Muelling, and Fragkiadaki]{agarwal2019model}
Arpit Agarwal, Katharina Muelling, and Katerina Fragkiadaki.
\newblock Model learning for look-ahead exploration in continuous control.
\newblock In \emph{Proceedings of the AAAI Conference on Artificial Intelligence}, pages 3151--3158, 2019.

\bibitem[An et~al.(2022)An, Wang, Li, Wang, Hong, Huang, Wang, and Shao]{an20221st}
Dong An, Zun Wang, Yangguang Li, Yi Wang, Yicong Hong, Yan Huang, Liang Wang, and Jing Shao.
\newblock 1st place solutions for rxr-habitat vision-and-language navigation competition (cvpr 2022).
\newblock \emph{arXiv preprint arXiv:2206.11610}, 2022.

\bibitem[An et~al.(2023{\natexlab{a}})An, Qi, Li, Huang, Wang, Tan, and Shao]{an2023bevbert}
Dong An, Yuankai Qi, Yangguang Li, Yan Huang, Liang Wang, Tieniu Tan, and Jing Shao.
\newblock Bevbert: Multimodal map pre-training for language-guided navigation.
\newblock In \emph{ICCV}, pages 2737--2748, 2023{\natexlab{a}}.

\bibitem[An et~al.(2023{\natexlab{b}})An, Wang, Wang, Wang, Huang, He, and Wang]{an2023etpnav}
Dong An, Hanqing Wang, Wenguan Wang, Zun Wang, Yan Huang, Keji He, and Liang Wang.
\newblock Etpnav: Evolving topological planning for vision-language navigation in continuous environments.
\newblock \emph{arXiv preprint arXiv:2304.03047}, 2023{\natexlab{b}}.

\bibitem[Anderson et~al.(2018)Anderson, Wu, Teney, Bruce, Johnson, S{\"u}nderhauf, Reid, Gould, and Van Den~Hengel]{VLN_2018vision}
Peter Anderson, Qi Wu, Damien Teney, Jake Bruce, Mark Johnson, Niko S{\"u}nderhauf, Ian Reid, Stephen Gould, and Anton Van Den~Hengel.
\newblock Vision-and-language navigation: Interpreting visually-grounded navigation instructions in real environments.
\newblock In \emph{CVPR}, pages 3674--3683, 2018.

\bibitem[Chang et~al.(2017)Chang, Dai, Funkhouser, Halber, Niessner, Savva, Song, Zeng, and Zhang]{matterport3d}
Angel Chang, Angela Dai, Thomas Funkhouser, Maciej Halber, Matthias Niessner, Manolis Savva, Shuran Song, Andy Zeng, and Yinda Zhang.
\newblock Matterport3d: Learning from rgb-d data in indoor environments.
\newblock In \emph{3DV}, pages 667--676, 2017.

\bibitem[Chen et~al.(2022{\natexlab{a}})Chen, Ji, Lin, Zeng, Li, Tan, and Gan]{chen2022weakly}
Peihao Chen, Dongyu Ji, Kunyang Lin, Runhao Zeng, Thomas~H Li, Mingkui Tan, and Chuang Gan.
\newblock Weakly-supervised multi-granularity map learning for vision-and-language navigation.
\newblock In \emph{NeurIPS}, 2022{\natexlab{a}}.

\bibitem[Chen et~al.(2021)Chen, Guhur, Schmid, and Laptev]{chen2021history}
Shizhe Chen, Pierre-Louis Guhur, Cordelia Schmid, and Ivan Laptev.
\newblock History aware multimodal transformer for vision-and-language navigation.
\newblock In \emph{NeurIPS}, pages 5834--5847, 2021.

\bibitem[Chen et~al.(2022{\natexlab{b}})Chen, Guhur, Tapaswi, Schmid, and Laptev]{chen2022think-GL}
Shizhe Chen, Pierre-Louis Guhur, Makarand Tapaswi, Cordelia Schmid, and Ivan Laptev.
\newblock Think global, act local: Dual-scale graph transformer for vision-and-language navigation.
\newblock In \emph{CVPR}, pages 16537--16547, 2022{\natexlab{b}}.

\bibitem[DeVries et~al.(2021)DeVries, Bautista, Srivastava, Taylor, and Susskind]{devries2021unconstrained}
Terrance DeVries, Miguel~Angel Bautista, Nitish Srivastava, Graham~W Taylor, and Joshua~M Susskind.
\newblock Unconstrained scene generation with locally conditioned radiance fields.
\newblock In \emph{ICCV}, pages 14304--14313, 2021.

\bibitem[Feng et~al.(2022)Feng, Fu, Lu, and Wang]{feng2022uln}
Weixi Feng, Tsu-Jui Fu, Yujie Lu, and William~Yang Wang.
\newblock Uln: Towards underspecified vision-and-language navigation.
\newblock \emph{arXiv preprint arXiv:2210.10020}, 2022.

\bibitem[Fried et~al.(2018)Fried, Hu, Cirik, Rohrbach, Andreas, Morency, Berg-Kirkpatrick, Saenko, Klein, and Darrell]{2018-speaker}
Daniel Fried, Ronghang Hu, Volkan Cirik, Anna Rohrbach, Jacob Andreas, Louis-Philippe Morency, Taylor Berg-Kirkpatrick, Kate Saenko, Dan Klein, and Trevor Darrell.
\newblock Speaker-follower models for vision-and-language navigation.
\newblock In \emph{NeurIPS}, 2018.

\bibitem[Georgakis et~al.(2022)Georgakis, Schmeckpeper, Wanchoo, Dan, Miltsakaki, Roth, and Daniilidis]{georgakis2022cm2}
Georgios Georgakis, Karl Schmeckpeper, Karan Wanchoo, Soham Dan, Eleni Miltsakaki, Dan Roth, and Kostas Daniilidis.
\newblock Cross-modal map learning for vision and language navigation.
\newblock In \emph{CVPR}, 2022.

\bibitem[Grandits et~al.(2021)Grandits, Effland, Pock, Krause, Plank, and Pezzuto]{grandits_geasi_2021}
Thomas Grandits, Alexander Effland, Thomas Pock, Rolf Krause, Gernot Plank, and Simone Pezzuto.
\newblock {GEASI}: {Geodesic}-based earliest activation sites identification in cardiac models.
\newblock \emph{International Journal for Numerical Methods in Biomedical Engineering}, 37\penalty0 (8):\penalty0 e3505, 2021.

\bibitem[Hong et~al.(2021)Hong, Wu, Qi, Rodriguez-Opazo, and Gould]{hong2021vln-bert}
Yicong Hong, Qi Wu, Yuankai Qi, Cristian Rodriguez-Opazo, and Stephen Gould.
\newblock Vln bert: A recurrent vision-and-language bert for navigation.
\newblock In \emph{CVPR}, pages 1643--1653, 2021.

\bibitem[Hong et~al.(2022)Hong, Wang, Wu, and Gould]{Hong2022bridging}
Yicong Hong, Zun Wang, Qi Wu, and Stephen Gould.
\newblock Bridging the gap between learning in discrete and continuous environments for vision-and-language navigation.
\newblock In \emph{CVPR}, 2022.

\bibitem[Hong et~al.(2023)Hong, Zhou, Zhang, Dernoncourt, Bui, Gould, and Tan]{hong2023learning}
Yicong Hong, Yang Zhou, Ruiyi Zhang, Franck Dernoncourt, Trung Bui, Stephen Gould, and Hao Tan.
\newblock Learning navigational visual representations with semantic map supervision.
\newblock In \emph{ICCV}, pages 3055--3067, 2023.

\bibitem[Huang et~al.(2023)Huang, Mees, Zeng, and Burgard]{huang23vlmaps}
Chenguang Huang, Oier Mees, Andy Zeng, and Wolfram Burgard.
\newblock Visual language maps for robot navigation.
\newblock In \emph{ICRA}, London, UK, 2023.

\bibitem[Ke et~al.(2019)Ke, Li, Bisk, Holtzman, Gan, Liu, Gao, Choi, and Srinivasa]{ke2019tactical}
Liyiming Ke, Xiujun Li, Yonatan Bisk, Ari Holtzman, Zhe Gan, Jingjing Liu, Jianfeng Gao, Yejin Choi, and Siddhartha Srinivasa.
\newblock Tactical rewind: Self-correction via backtracking in vision-and-language navigation.
\newblock In \emph{Proceedings of the IEEE/CVF conference on computer vision and pattern recognition}, pages 6741--6749, 2019.

\bibitem[Kerbl et~al.(2023)Kerbl, Kopanas, Leimk{\"u}hler, and Drettakis]{kerbl20233d}
Bernhard Kerbl, Georgios Kopanas, Thomas Leimk{\"u}hler, and George Drettakis.
\newblock 3d gaussian splatting for real-time radiance field rendering.
\newblock \emph{ACM Transactions on Graphics}, 42\penalty0 (4):\penalty0 1--14, 2023.

\bibitem[Koh et~al.(2021)Koh, Lee, Yang, Baldridge, and Anderson]{koh2021pathdreamer}
Jing~Yu Koh, Honglak Lee, Yinfei Yang, Jason Baldridge, and Peter Anderson.
\newblock Pathdreamer: A world model for indoor navigation.
\newblock In \emph{ICCV}, pages 14738--14748, 2021.

\bibitem[Krantz and Lee(2022)]{krantz2022sim2sim}
Jacob Krantz and Stefan Lee.
\newblock Sim-2-sim transfer for vision-and-language navigation in continuous environments.
\newblock In \emph{ECCV}, 2022.

\bibitem[Krantz et~al.(2020)Krantz, Wijmans, Majumdar, Batra, and Lee]{Krantz2020r2r-ce}
Jacob Krantz, Erik Wijmans, Arjun Majumdar, Dhruv Batra, and Stefan Lee.
\newblock Beyond the nav-graph: Vision-and-language navigation in continuous environments.
\newblock In \emph{ECCV}, 2020.

\bibitem[Ku et~al.(2020)Ku, Anderson, Patel, Ie, and Baldridge]{2020_RXR}
Alexander Ku, Peter Anderson, Roma Patel, Eugene Ie, and Jason Baldridge.
\newblock Room-across-room: Multilingual vision-and-language navigation with dense spatiotemporal grounding.
\newblock In \emph{EMNLP}, pages 4392--4412, 2020.

\bibitem[Kwon et~al.(2023)Kwon, Park, and Oh]{kwon2023renderable}
Obin Kwon, Jeongho Park, and Songhwai Oh.
\newblock Renderable neural radiance map for visual navigation.
\newblock In \emph{CVPR}, pages 9099--9108, 2023.

\bibitem[Li and Bansal(2023)]{li2023improving}
Jialu Li and Mohit Bansal.
\newblock Improving vision-and-language navigation by generating future-view image semantics.
\newblock In \emph{CVPR}, pages 10803--10812, 2023.

\bibitem[Li et~al.(2023)Li, Wang, Yang, Wang, and Jiang]{Li2023KERM}
Xiangyang Li, Zihan Wang, Jiahao Yang, Yaowei Wang, and Shuqiang Jiang.
\newblock Kerm: Knowledge enhanced reasoning for vision-and-language navigation.
\newblock In \emph{CVPR}, pages 2583--2592, 2023.

\bibitem[Liu et~al.(2023)Liu, Wang, Wang, and Yang]{liu2023bird}
Rui Liu, Xiaohan Wang, Wenguan Wang, and Yi Yang.
\newblock Bird's-eye-view scene graph for vision-language navigation.
\newblock In \emph{ICCV}, pages 10968--10980, 2023.

\bibitem[Mildenhall et~al.(2021)Mildenhall, Srinivasan, Tancik, Barron, Ramamoorthi, and Ng]{mildenhall2021nerf}
Ben Mildenhall, Pratul~P Srinivasan, Matthew Tancik, Jonathan~T Barron, Ravi Ramamoorthi, and Ren Ng.
\newblock Nerf: Representing scenes as neural radiance fields for view synthesis.
\newblock \emph{Communications of the ACM}, 65\penalty0 (1):\penalty0 99--106, 2021.

\bibitem[Pashevich et~al.(2021)Pashevich, Schmid, and Sun]{pashevich2021episodic}
Alexander Pashevich, Cordelia Schmid, and Chen Sun.
\newblock Episodic transformer for vision-and-language navigation.
\newblock In \emph{ICCV}, 2021.

\bibitem[Pedregosa et~al.(2011)Pedregosa, Varoquaux, Gramfort, Michel, Thirion, Grisel, Blondel, Prettenhofer, Weiss, Dubourg, et~al.]{pedregosa2011scikit}
Fabian Pedregosa, Ga{\"e}l Varoquaux, Alexandre Gramfort, Vincent Michel, Bertrand Thirion, Olivier Grisel, Mathieu Blondel, Peter Prettenhofer, Ron Weiss, Vincent Dubourg, et~al.
\newblock Scikit-learn: Machine learning in python.
\newblock \emph{Journal of machine Learning research}, 12:\penalty0 2825--2830, 2011.

\bibitem[Qi et~al.(2020)Qi, Wu, Anderson, Wang, Wang, Shen, and Hengel]{2020reverie}
Yuankai Qi, Qi Wu, Peter Anderson, Xin Wang, William~Yang Wang, Chunhua Shen, and Anton van~den Hengel.
\newblock Reverie: Remote embodied visual referring expression in real indoor environments.
\newblock In \emph{CVPR}, pages 9982--9991, 2020.

\bibitem[Radford et~al.(2021)Radford, Kim, Hallacy, Ramesh, Goh, Agarwal, Sastry, Askell, Mishkin, Clark, et~al.]{radford2021learning}
Alec Radford, Jong~Wook Kim, Chris Hallacy, Aditya Ramesh, Gabriel Goh, Sandhini Agarwal, Girish Sastry, Amanda Askell, Pamela Mishkin, Jack Clark, et~al.
\newblock Learning transferable visual models from natural language supervision.
\newblock In \emph{ICML}, pages 8748--8763, 2021.

\bibitem[Ramakrishnan et~al.(2021)Ramakrishnan, Gokaslan, Wijmans, Maksymets, Clegg, Turner, Undersander, Galuba, Westbury, Chang, et~al.]{ramakrishnan2021habitat}
Santhosh~K Ramakrishnan, Aaron Gokaslan, Erik Wijmans, Oleksandr Maksymets, Alex Clegg, John Turner, Eric Undersander, Wojciech Galuba, Andrew Westbury, Angel~X Chang, et~al.
\newblock Habitat-matterport 3d dataset (hm3d): 1000 large-scale 3d environments for embodied ai.
\newblock \emph{arXiv preprint arXiv:2109.08238}, 2021.

\bibitem[Ramesh et~al.(2021)Ramesh, Pavlov, Goh, Gray, Voss, Radford, Chen, and Sutskever]{ramesh2021zero}
Aditya Ramesh, Mikhail Pavlov, Gabriel Goh, Scott Gray, Chelsea Voss, Alec Radford, Mark Chen, and Ilya Sutskever.
\newblock Zero-shot text-to-image generation.
\newblock In \emph{ICML}, pages 8821--8831. PMLR, 2021.

\bibitem[Saharia et~al.(2022)Saharia, Chan, Chang, Lee, Ho, Salimans, Fleet, and Norouzi]{saharia2022palette}
Chitwan Saharia, William Chan, Huiwen Chang, Chris Lee, Jonathan Ho, Tim Salimans, David Fleet, and Mohammad Norouzi.
\newblock Palette: Image-to-image diffusion models.
\newblock In \emph{ACM SIGGRAPH 2022 conference proceedings}, pages 1--10, 2022.

\bibitem[Shen et~al.(2023)Shen, Yang, Yu, Wong, Kaelbling, and Isola]{pmlr-v229-shen23a}
William Shen, Ge Yang, Alan Yu, Jansen Wong, Leslie~Pack Kaelbling, and Phillip Isola.
\newblock Distilled feature fields enable few-shot language-guided manipulation.
\newblock In \emph{Proceedings of The 7th Conference on Robot Learning}, pages 405--424, 2023.

\bibitem[Taioli et~al.(2023)Taioli, Cunico, Girella, Bologna, Farinelli, and Cristani]{taioli2023language}
Francesco Taioli, Federico Cunico, Federico Girella, Riccardo Bologna, Alessandro Farinelli, and Marco Cristani.
\newblock Language-enhanced rnr-map: Querying renderable neural radiance field maps with natural language.
\newblock In \emph{ICCV}, pages 4669--4674, 2023.

\bibitem[Tan et~al.(2019)Tan, Yu, and Bansal]{tan2019learning}
Hao Tan, Licheng Yu, and Mohit Bansal.
\newblock Learning to navigate unseen environments: Back translation with environmental dropout.
\newblock In \emph{NAACL}, pages 2610--2621, 2019.

\bibitem[Thomason et~al.(2020)Thomason, Murray, Cakmak, and Zettlemoyer]{thomason2020cvdn}
Jesse Thomason, Michael Murray, Maya Cakmak, and Luke Zettlemoyer.
\newblock Vision-and-dialog navigation.
\newblock In \emph{PMLR}, 2020.

\bibitem[Wang et~al.(2020)Wang, Wang, Shu, Liang, and Shen]{wang2020active}
Hanqing Wang, Wenguan Wang, Tianmin Shu, Wei Liang, and Jianbing Shen.
\newblock Active visual information gathering for vision-language navigation.
\newblock In \emph{Computer Vision--ECCV 2020: 16th European Conference, Glasgow, UK, August 23--28, 2020, Proceedings, Part XXII 16}, pages 307--322. Springer, 2020.

\bibitem[Wang et~al.(2023{\natexlab{a}})Wang, Liang, Van~Gool, and Wang]{wang2023dreamwalker}
Hanqing Wang, Wei Liang, Luc Van~Gool, and Wenguan Wang.
\newblock Dreamwalker: Mental planning for continuous vision-language navigation.
\newblock In \emph{ICCV}, pages 10873--10883, 2023{\natexlab{a}}.

\bibitem[Wang et~al.(2019)Wang, Huang, Celikyilmaz, Gao, Shen, Wang, Wang, and Zhang]{2019reinforced}
Xin Wang, Qiuyuan Huang, Asli Celikyilmaz, Jianfeng Gao, Dinghan Shen, Yuan-Fang Wang, William~Yang Wang, and Lei Zhang.
\newblock Reinforced cross-modal matching and self-supervised imitation learning for vision-language navigation.
\newblock In \emph{CVPR}, pages 6629--6638, 2019.

\bibitem[Wang et~al.(2023{\natexlab{b}})Wang, Liu, Song, Wang, and Jiang]{wang2023generating}
Xiaohan Wang, Yuehu Liu, Xinhang Song, Beibei Wang, and Shuqiang Jiang.
\newblock Generating explanations for embodied action decision from visual observation.
\newblock In \emph{Proceedings of the 31st ACM International Conference on Multimedia}, pages 2838--2846, 2023{\natexlab{b}}.

\bibitem[Wang et~al.(2024)Wang, Liu, Song, Wang, and Jiang]{wang2024camp}
Xiaohan Wang, Yuehu Liu, Xinhang Song, Beibei Wang, and Shuqiang Jiang.
\newblock Camp: Causal multi-policy planning for interactive navigation in multi-room scenes.
\newblock \emph{Advances in Neural Information Processing Systems}, 36, 2024.

\bibitem[Wang et~al.(2023{\natexlab{c}})Wang, Li, Hong, Wang, Wu, Bansal, Gould, Tan, and Qiao]{wang2023scaling}
Zun Wang, Jialu Li, Yicong Hong, Yi Wang, Qi Wu, Mohit Bansal, Stephen Gould, Hao Tan, and Yu Qiao.
\newblock Scaling data generation in vision-and-language navigation.
\newblock In \emph{ICCV}, pages 12009--12020, 2023{\natexlab{c}}.

\bibitem[Wang et~al.(2023{\natexlab{d}})Wang, Li, Yang, Liu, and Jiang]{wang2023gridmm}
Zihan Wang, Xiangyang Li, Jiahao Yang, Yeqi Liu, and Shuqiang Jiang.
\newblock Gridmm: Grid memory map for vision-and-language navigation.
\newblock In \emph{ICCV}, pages 15625--15636, 2023{\natexlab{d}}.

\bibitem[Zhang et~al.(2021)Zhang, Song, Bai, Li, Chu, and Jiang]{zsx_ICCV21}
Sixian Zhang, Xinhang Song, Yubing Bai, Weijie Li, Yakui Chu, and Shuqiang Jiang.
\newblock Hierarchical object-to-zone graph for object navigation.
\newblock In \emph{2021 {IEEE/CVF} International Conference on Computer Vision, {ICCV} 2021, Montreal, QC, Canada, October 10-17, 2021}, pages 15110--15120. {IEEE}, 2021.

\bibitem[Zhang et~al.(2022)Zhang, Li, Song, Bai, and Jiang]{zsx_ECCV22}
Sixian Zhang, Weijie Li, Xinhang Song, Yubing Bai, and Shuqiang Jiang.
\newblock Generative meta-adversarial network for unseen object navigation.
\newblock In \emph{Computer Vision - {ECCV} 2022 - 17th European Conference, Tel Aviv, Israel, October 23-27, 2022, Proceedings, Part {XXXIX}}, pages 301--320, 2022.

\bibitem[Zhang et~al.(2023)Zhang, Song, Li, Bai, Yu, and Jiang]{zsx_CVPR23}
Sixian Zhang, Xinhang Song, Weijie Li, Yubing Bai, Xinyao Yu, and Shuqiang Jiang.
\newblock Layout-based causal inference for object navigation.
\newblock In \emph{Proceedings of the IEEE/CVF Conference on Computer Vision and Pattern Recognition (CVPR)}, pages 10792--10802, 2023.

\bibitem[Zhu et~al.(2021)Zhu, Liang, Zhu, Yu, Chang, and Liang]{zhu2021soon}
Fengda Zhu, Xiwen Liang, Yi Zhu, Qizhi Yu, Xiaojun Chang, and Xiaodan Liang.
\newblock Soon: Scenario oriented object navigation with graph-based exploration.
\newblock In \emph{CVPR}, pages 12689--12699, 2021.

\end{thebibliography}
